\documentclass{ifacconf}
\makeatletter
\def\ps@plain{\let\@oddfoot\@empty}
\makeatother
\usepackage{graphicx}      
\usepackage{natbib}     
\usepackage{algorithm}
\usepackage{graphicx}      
\usepackage{amsmath}      
\usepackage{amssymb}       
\usepackage{algorithm}     
\usepackage{algorithmic}   
\usepackage{booktabs}      
\usepackage{multirow}      
\usepackage{subcaption}    
\usepackage{url}  
\usepackage{xcolor}
\usepackage{booktabs}
\usepackage{array}
\usepackage{graphicx}
\usepackage{xcolor}
\usepackage{colortbl}
\usepackage{soul}
\usepackage[most]{tcolorbox}
\soulregister\cite7
\soulregister\ref7
\soulregister\eqref7
\soulregister\emph1

\definecolor{revone}{RGB}{255,240,220}
\definecolor{revonebg}{RGB}{255,240,220}

\definecolor{revtwo}{RGB}{230,255,230}
\definecolor{revtwobg}{RGB}{230,255,230}

\definecolor{revthree}{RGB}{230,240,255}
\definecolor{revthreebg}{RGB}{230,240,255}
\newcommand{\revthreehl}[1]{{\sethlcolor{revthree}\hl{#1}}}

\definecolor{editor}{RGB}{180,180,180}
\definecolor{editorbg}{RGB}{245,245,245}

\definecolor{editor}{RGB}{195,95,230}
\definecolor{editorbg}{RGB}{250,242,255}

\definecolor{revonemarkbg}{RGB}{255,145,200}

\definecolor{revtwomarkbg}{RGB}{200,240,240}

\definecolor{revthreemarkbg}{RGB}{230,210,250}

\newcounter{editorcmtcnt}
\newcounter{revonecmtcnt}
\newcounter{revtwocmtcnt}
\newcounter{revthreecmtcnt}

\newtcolorbox{editorcmt}[1][]{
  enhanced,
  title={\refstepcounter{editorcmtcnt}Comment \theeditorcmtcnt\ifx&#1&\else: #1\fi},
  colframe=editor,
  colback=editorbg,
  colbacktitle=editor,
  coltitle=black,
  fonttitle=\bfseries,
  boxrule=0.5pt,
  arc=1pt,
  toptitle=2pt, bottomtitle=2pt,
  left=6pt, right=6pt, top=6pt, bottom=6pt
}

\newtcolorbox{revonecmt}[1][]{
  enhanced,
  title={\refstepcounter{revonecmtcnt}Comment 1.\therevonecmtcnt\ifx&#1&\else: #1\fi},
  colframe=revone,
  colback=revonebg,
  colbacktitle=revone,
  coltitle=black,
  fonttitle=\bfseries,
  boxrule=0.5pt,
  arc=1pt,
  toptitle=2pt, bottomtitle=2pt,
  left=6pt, right=6pt, top=6pt, bottom=6pt
}

\newtcolorbox{revtwocmt}[1][]{
  enhanced,
  title={\refstepcounter{revtwocmtcnt}Comment 2.\therevtwocmtcnt\ifx&#1&\else: #1\fi},
  colframe=revtwo,
  colback=revtwobg,
  colbacktitle=revtwo,
  coltitle=black,
  fonttitle=\bfseries,
  boxrule=0.5pt,
  arc=1pt,
  toptitle=2pt, bottomtitle=2pt,
  left=6pt, right=6pt, top=6pt, bottom=6pt
}

\newtcolorbox{revthreecmt}[1][]{
  enhanced,
  title={\refstepcounter{revthreecmtcnt}Comment 3.\therevthreecmtcnt\ifx&#1&\else: #1\fi},
  colframe=revthree,
  colback=revthreebg,
  colbacktitle=revthree,
  coltitle=black,
  fonttitle=\bfseries,
  boxrule=0.5pt,
  arc=1pt,
  toptitle=2pt, bottomtitle=2pt,
  left=6pt, right=6pt, top=6pt, bottom=6pt
}

\begin{document}
\begin{frontmatter}

\makeatletter
\def\title@fmt#1#2{%
  \@ifundefined{@runtitle}{\global\def\@runtitle{#1}}{}%
  \@articletypesize
  \leavevmode\vphantom{Aye!}
  \@articletype
  \vskip12\p@
  {\@titlesize
    \noindent\hspace*{-10.5mm}%
    \parbox{0.7\textwidth}{\centering #1\\$^{#2}$}%
    \hfil
  }%
  \vskip\@undertitleskip
}
\makeatother

\title{Generalist AI Control: Towards Multi-purpose Adaptive Algorithms} 

\author[First]{Klinsmann Agyei, Pouria Sarhadi} 

\address[First]{School of Physics, Engineering and Computer Science, University of Hertfordshire, 
   Hatfield, UK (e-mail: \textcolor{blue}{k.agyei@herts.ac.uk}, \textcolor{blue}{p.sarhadi@herts.ac.uk})}

\begin{abstract}
Traditional controllers are designed for specific systems and do not transfer across different system orders and dynamics. We present a Generalist Controller, a learning-based controller capable of controlling systems of varying orders and dynamics. The approach introduces a novel dynamic state-space representation using attention mechanisms with masking, enabling a single neural network, trained in one shot, to handle systems with different dimensions without architectural modifications by assigning a system tag to each system. We generated 314,630 demonstrations from 25 diverse systems, including stable, unstable, minimum-phase, and non-minimum-phase dynamics, spanning linear and nonlinear systems from autonomous underwater and aerospace vehicles to mechanical systems and chemical processes. The model learns cross-system control strategies through multi-scale temporal processing and a mixture-of-experts architecture. Simulation results demonstrate that the proposed generalist controller achieves comparable performance to system-specific LQI controllers across all tested systems, including challenging cases such as non-minimum-phase and unstable dynamics, whilst generalising to unseen operating conditions including actuator saturation, noise, disturbance, and reference trajectories not encountered during training. This work represents a significant step towards generalist control policies within a defined family of dynamical systems, demonstrating effective control across a range of single-input single-output (SISO) systems of varying order and dynamics using a single learned policy without system-specific tuning.

\end{abstract}


\end{frontmatter}

\begin{center}
\begin{minipage}{0.95\linewidth}
\footnotesize
\textit{This is the accepted manuscript of the article published in
\emph{Control Engineering Practice}. The final published version is
available at:
\url{https://doi.org/10.1016/j.conengprac.2026.106915}.}

\medskip

\textcopyright\ 2026 The Authors. This work is licensed under the
Creative Commons Attribution 4.0 International License:
\url{https://creativecommons.org/licenses/by/4.0/}.
\end{minipage}
\end{center}

\section{Introduction}
\vspace{-10pt}
The control engineering discipline stands at an inflexion point where traditional model-based design paradigms are being complemented by data-driven approaches. Classical control methodologies \cite{mayne2000constrained, astrom1994adaptive, doyle1996robust} have proven effective across countless applications that shape our daily lives. 

However, these approaches require accurate system models and separate controller design for each deployment. For many practical systems or systems with poorly characterised dynamics, accurate models are difficult or expensive to obtain and maintain. This motivates learning-based controllers that acquire control behaviour directly from demonstrations rather than analytical models. Whilst such approaches are well established \cite{pomerleau1988alvinn, mandlekar2021matters,argall2009survey,schaal1999imitation}, they typically train separate policies for each system. We propose a generalist learning-based controller, a single policy trained on demonstrations from diverse systems without per-system tuning. Crucially, the framework is agnostic to the demonstration source, it can learn from model-based experts where accurate models exist, or from human teleoperation and recorded operational data where they do not, unifying both scenarios under a single deployable policy.

The transformative impact of foundation models across artificial intelligence domains has sparked considerable interest in their application to control systems. In natural language processing, models such as \cite{brown2020language} have demonstrated remarkable few-shot generalisation capabilities through pre-training on diverse textual corpora, with applications on higher-level decision making in autonomous vehicles \cite{agyei2025large}. Most pertinently for our domain, the robotics community has witnessed the emergence of foundation models specifically targeting embodied intelligence, suggesting that similar paradigms might revolutionise control theory itself. \cite{octo2024}, a transformer-based policy trained on 800k robot trajectories across diverse embodiments, has shown that cross-robot generalisation is achievable through large-scale pre-training. Before this, \cite{reed2022generalist}, DeepMind's multi-modal, multi-task, multi-embodiment generalist agent, controls multiple robots whilst simultaneously performing language tasks and playing Atari games with a single set of neural weights. \cite{brohan2022rt} and \cite{zitkovich2023rt} have achieved zero-shot generalisation to new tasks and environments, bridging the semantic gap between high-level instructions and low-level control through vision-language-action models. NVIDIA's recently announced GR00T N1 \cite{bjorck2025gr00t} represents a generalist model for humanoid robots, featuring a dual-system architecture inspired by human cognition that enables generalised reasoning and manipulation skills across different embodiments.

However, these models fundamentally operate at the perception-action level rather than addressing the underlying control-theoretic challenges. Models like \cite{octo2024} and \cite{reed2022generalist} learn control policies jointly with visual representations, creating entangled representations where control decisions may become linked to visual features. This visual-motor coupling could introduce some fundamental limitations. These models may be less effective without camera inputs, rendering them less suitable for applied control problems.

Parallel to these developments, imitation learning and behavioural cloning \cite{pomerleau1988alvinn,bratko1995BCIL,Zare2024BCIM} have emerged as powerful paradigms for acquiring control policies from demonstrations. These approaches offer several advantages over model-based control design, they can capture complex control strategies that are difficult to derive analytically, they do not require explicit system models at deployment, and they can leverage existing expert controllers or human demonstrations. However, most imitation learning approaches train separate policies for each system, failing to exploit the shared structure underlying diverse dynamical systems.

This paper introduces the Generalist Controller, a learned controller that addresses this fundamental gap by learning control policies that transfer across systems of varying dimensionality and dynamics. Unlike existing approaches that focus on task-level or perception-level generalisation, the proposed generalist controller operates at the control-theoretic level, learning to track references for systems of varying orders. The key insight underpinning this approach is that control at the state-space level represents a highly transferable form of control knowledge across many physical systems. Whilst visual features vary dramatically across robots and environments, the underlying mathematics of dynamical systems remain invariant. A mass-spring-damper system exhibits identical second-order dynamics whether implemented as a vehicle suspension, a building's seismic isolator, or a microscale MEMS device. By learning control policies directly in state-space, we capture these fundamental dynamical patterns that transcend specific embodiments or sensory modalities. This approach introduces several technical innovations to enable cross-order generalisation. We develop a dynamic state-space representation with attention masking that allows a single neural network to process systems of different orders through padding. Through multi-scale temporal processing and mixture-of-experts architecture \cite{shazeer2017outrageously}, the controller learns to recognise and respond to different dynamical regimes, from stable first-order responses to complex oscillatory behaviours. Trained on 314,630 expert demonstrations from 25 diverse systems, the proposed generalist controller achieves performance comparable to system-specific Linear Quadratic Integral (LQI) controllers. While the broader motivation of this work is inspired by recent foundation model approaches, the scope of the present study is more specific. In this paper, we focus on SISO dynamical systems.

The implications of this work extend beyond technical achievements in conventional control and, more specifically, in imitation learning for control applications. By demonstrating that control knowledge transfers across diverse system dynamics within pre-trained system families, and is not limited to a single category of systems, we enable multi-system control policies that potentially require no per-system re-tuning. This represents a paradigm shift from the traditional approach of designing controllers for specific systems towards learning controllers that embody general principles applicable across entire system classes. As industries face increasing demands for flexible automation, rapid reconfiguration, and robust performance across varying operating conditions, generalist controllers offer a compelling alternative to traditional design methodologies.

Before formalising the problem, we clarify the use of the term \emph{generalist} as employed in this paper. The term \emph{generalist} is used in a pragmatic sense consistent with recent learning-based robotics and foundation-model literature, where a single parameterised policy is trained on data from multiple tasks or systems \cite{reed2022generalist, octo2024}. Adapting this notion to control engineering, we define a generalist controller as a single control policy trained on demonstrations from multiple dynamical systems and deployed without per-system retuning. The novelty of this work does not lie in introducing a new learning paradigm, as similar ideas have long been considered in Behavioural Cloning (BC) and Imitation Learning (IL) \cite{pomerleau1988alvinn,bratko1995BCIL,Zare2024BCIM}. Rather, leveraging recent advances in deep neural networks, we demonstrate that a single policy can effectively reuse control structure across heterogeneous systems with varying orders and dynamics, without any system-specific tuning. It should be noted that the controller requires a system tag as input to indicate which system is being controlled. It does not infer the system identity from observations alone and cannot generalise to systems not included in the training set. This parallels human adaptive behaviour, which can control systems after training and does not start blindly when encountering different systems.

The remainder of the paper is organised as follows. Section~2 reviews related work in adaptive, robust, and generalisation in control. Section~3 formulates the problem and defines the learning objective. Section~4 presents the proposed Generalist Controller architecture. Section~5 describes the training framework and data generation. Section~6 reports simulation and hardware results, including comparisons with classical LQI controllers. Section~7 presents ablation studies analysing, the role of key architectural components. Section~8 concludes the paper and discusses limitations and future directions.

\vspace{-0.3cm}
\section{Related Work}
\vspace{-10pt}
This section reviews prior approaches to achieving control across multiple systems, spanning classical adaptive and robust control, generalisation methods in control theory, and recent learning-based approaches. We highlight the limitations of existing methods in handling systems with varying orders and dynamic characteristics, motivating our approach of learning a single policy from demonstrations across diverse systems.

\subsection{Classical Adaptive and Robust Control Theory}
\vspace{-10pt}
Classical adaptive control provides theoretical frameworks for handling uncertainty, but focuses on adapting to parameter variations within a fixed system structure. Model Reference Adaptive Control (MRAC) \cite{ioannou2006adaptive} adjusts controller parameters to match reference model behaviour for a specific plant. Self-tuning regulators \cite{astrom1994adaptive} estimate system parameters online and update control laws accordingly. Neural adaptive control \cite{lewis2002neural} combines neural networks with Lyapunov-based adaptation laws. Whilst effective for parametric uncertainty, these methods assume a known system order and structure, requiring redesign when applied to systems with different dynamics.

Robust control addresses uncertainty through worst-case design. $H_\infty$ control \cite{zhou1996robust} minimises the worst-case gain from disturbances to outputs. $\mu$-synthesis \cite{doyle1982analysis} handles structured uncertainty. Whilst providing strong theoretical guarantees, robust controllers are conservative and require bounds on uncertainty. Recent work on learning-based robust control \cite{berkenkamp2017safe, richards2018lyapunov} combines data-driven methods with robust control theory, but remains limited to specific system classes. Simultaneous stabilisation \cite{vidyasagar2022similtaneouscontrol} represented another promising approach to address the generalisability issues of conventional controllers. The aim was to synthesise a single controller capable of stabilising multiple plants. However, these controllers are limited to plants that share specific structural properties, such as possessing the same number of unstable poles or satisfying parity interlacing conditions. In practice, these theoretical limitations also translate into significant engineering effort, as classical adaptive and robust controllers typically require careful, system-specific modelling, synthesis, and tuning for each new plant.

This work addresses these limitations by learning a single control policy that operates across systems with different structures and orders.

\subsection{Generalisation in Control Systems}
\vspace{-10pt}
The challenge of generalisation in control extends beyond parameter uncertainty to encompass structural variations, changing objectives, and diverse operating conditions. Koopman operator theory \cite{brunton2022data} provides a theoretical framework for lifting nonlinear dynamics into infinite-dimensional linear representations, potentially enabling transfer across system classes. However, practical implementations require finite-dimensional approximations that may not preserve important dynamical properties. Behavioural systems theory \cite{willems2007behavioral} offers an alternative perspective, focusing on trajectories rather than state-space representations. Whilst this approach provides elegant theoretical insights, practical controller synthesis within this framework remains computationally challenging for complex systems. This work maintains the computational tractability of state-space methods whilst achieving the generalisation benefits sought by behavioural approaches.

Recent advances in reinforcement learning have demonstrated impressive generalisation in game-playing and simulated environments \cite{schrittwieser2020mastering}, yet transfer to physical systems remains challenging. The reality gap, encompassing differences in dynamics, sensing, and actuation between simulation and reality, continues to limit practical deployment \cite{zhao2020sim}. Sim-to-real transfer techniques, including domain randomisation \cite{Tobin2017DomainRF} and system identification \cite{torricelli2018eurobench}, partially address these challenges but require extensive engineering effort for each new system. Our approach sidesteps many of these issues by operating directly on state-space representations, avoiding the complexity of visual sim-to-real transfer whilst maintaining the benefits of learning-based generalisation.

\section{Problem Formulation}
\vspace{-10pt}
We consider a collection of $N$ dynamical systems, denoted by $\mathcal{S}_i$, where each system may be linear or nonlinear, stable or unstable, and may exhibit minimum or non-minimum-phase behaviour. Each system $\mathcal{S}_i$ is characterised by its state $\mathbf{x}_{pi}(t) \in \mathbb{R}^{n_{x_p,i}}$, control input $\mathbf{u}_{pi}(t) \in \mathbb{R}^{n_{u_p,i}}$, and output $\mathbf{y}_{pi}(t) \in \mathbb{R}^{n_{y_p,i}}$.

For linear time-invariant (LTI) systems, the dynamics are described by
\begin{align}
\dot{\mathbf{x}}_{p_i}(t) &= A_{p_i} \mathbf{x_p}_{i}(t) + B_{p_i} \mathbf{u_p}_{i}(t), \label{eq:state_dynamics}\\
\mathbf{y}_{p_i}(t) &= C_{p_i} \mathbf{x}_{p_i}(t), \label{eq:output}
\end{align}
where $A_{p_i} \in \mathbb{R}^{n_{x_p,i} \times n_{x_p,i}}$, $B_{p_i} \in \mathbb{R}^{n_{x_p,i} \times n_{u_p,i}}$, and $C_{p_i} \in \mathbb{R}^{n_{y_p,i} \times n_{x_p,i}}$ are system-specific matrices.

For nonlinear systems, such as the REMUS Autonomous underwater vehicle (AUV) dynamics considered in this work, the system evolution is given by
\begin{align}
\dot{\mathbf{x}}_{p_i}(t) &= \mathbf{f_p}_i(\mathbf{x}_{p_i}(t), \mathbf{u}_{p_i}(t)), \label{eq:state_dynamics_nl}\\
\mathbf{y}_{p_i}(t) &= \mathbf{h_p}_{i}(\mathbf{x}_{p_i}(t)), \label{eq:output_nl}
\end{align}
where $\mathbf{f}_i: \mathbb{R}^{n_{x_p,i}} \times \mathbb{R}^{n_{u_p,i}} \rightarrow \mathbb{R}^{n_{x_p,i}}$ and $\mathbf{h}_{p_i}: \mathbb{R}^{n_{x_p,i}} \rightarrow \mathbb{R}^{n_{y_p,i}}$ denote the nonlinear state transition and output mappings.

The objective is to learn a single shared control policy $\pi_\theta$ that can be deployed across all systems $\mathcal{S}_i$:
\begin{equation}
 \pi_\theta:\; (\mathbf{x_p}_{i,t-T_h+1:t}, \mathbf{r}_{t-T_h+1:t}, \boldsymbol{\phi}_i)
\;\mapsto\;
\mathbf{u_p}_{i,t:t+H-1},
\label{eq:policy}
\end{equation}

where $T_h$ denotes the history window, $H$ is the prediction horizon, and $\theta$ are the shared parameters of the Generalist Controller. The policy receives a system tag or identifier as input, denoted by $\phi_i$, which is simply a labelled number assigned to each system. This label is assumed to be known at deployment. We emphasise that the controller does not perform implicit system identification or zero-shot adaptation to unknown dynamics. The need for this tag is similar to post-training human operation, where control is typically executed with awareness of which system is being controlled. This formulation enables receding-horizon control, allowing a single policy to be deployed across known heterogeneous linear and nonlinear systems without per-system tuning.

\section{The Proposed Generalist Controller}
\begin{figure*}[t]
\centering
\includegraphics[width=0.75\textwidth]{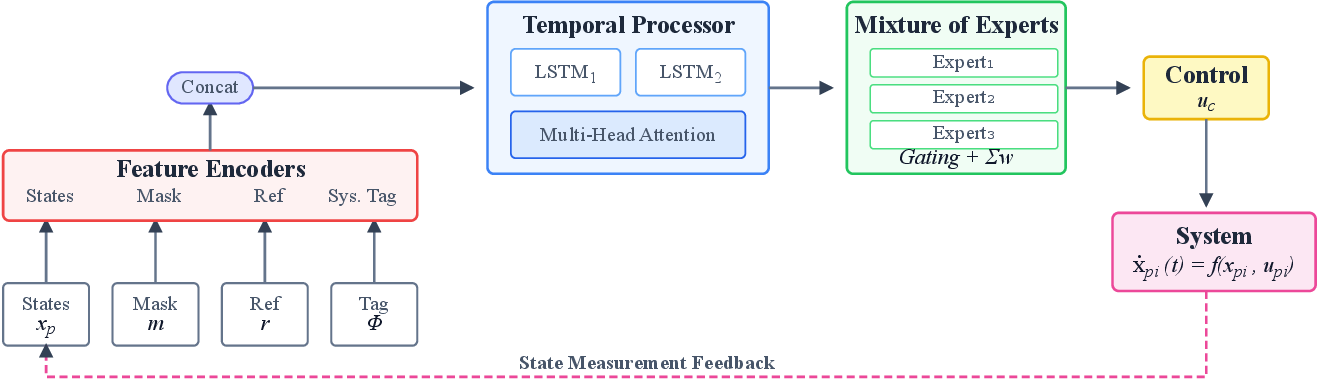}
\caption{Architecture of the proposed Generalist Controller}
\label{fig:architecture}
\end{figure*}

\vspace{-10pt}
The Generalist Controller employs a hierarchical architecture that progressively transforms raw state information into control actions. Figure~\ref{fig:architecture} illustrates the complete data flow: feature encoders process states with validity masking, references, and system parameters; temporal processors capture dynamics through parallel LSTMs and attention; and a mixture of experts generates control actions that are applied to the system.

For clarity, we summarise the operation of the proposed controller. At each timestep, the policy receives a history of measured states, the corresponding reference trajectory, and a discrete system tag that tells the controller which system it is controlling so it can select the appropriate control strategy for that specific system. State histories are masked to allow a unified representation across systems of different dimensions. Encoded features are processed by parallel temporal modules and mapped to control actions via a mixture-of-experts module. During training, the policy is optimised by supervised learning of LQI control actions, while at deployment it operates directly on measured states without online optimisation or per-system retuning.

\subsection{Feature Encoding}
The feature encoding stage processes four input streams to create a unified representation through parallel encoding networks.

The first two encoders process state vectors with validity masking to handle variable-order systems. Given the zero-padded state history $\mathbf{X} \in \mathbb{R}^{T_h \times n_{\max}}$ and validity mask $\mathbf{M} \in \{0,1\}^{T_h \times n_{\max}}$, the masked state is computed as $\tilde{\mathbf{X}} = \mathbf{X} \odot \mathbf{M}$, where $\odot$ denotes the Hadamard product. The state encoders map this to latent representations:
\begin{align}
\mathbf{f}_1 &= \mathbf{W}_2^{(1)} \, \sigma\!\left( \mathbf{W}_1^{(1)} \, \text{vec}(\tilde{\mathbf{X}}) + \mathbf{b}_1^{(1)} \right) + \mathbf{b}_2^{(1)} \\[4pt]
\mathbf{f}_2 &= \mathbf{W}_2^{(2)} \, \sigma\!\left( \mathbf{W}_1^{(2)} \, \text{vec}(\tilde{\mathbf{X}}) + \mathbf{b}_1^{(2)} \right) + \mathbf{b}_2^{(2)}
\end{align}
where $\sigma(\cdot)$ denotes the ReLU activation function, $\text{vec}(\cdot)$ vectorises the matrix input, and $\{\mathbf{W}_j^{(i)}, \mathbf{b}_j^{(i)}\}$ are learnable parameters for encoder $i$ and layer $j$.
The third encoder processes the reference trajectory $\mathbf{r} \in \mathbb{R}^{T_h}$, whilst the fourth encoder processes the system parameter vector $\boldsymbol{\phi} \in \mathbb{R}^{n_\phi}$:
\begin{align}
\mathbf{f}_3 &= \mathbf{W}_2^{(3)} \, \sigma\!\left( \mathbf{W}_1^{(3)} \mathbf{r} + \mathbf{b}_1^{(3)} \right) + \mathbf{b}_2^{(3)} \\[4pt]
\mathbf{f}_4 &= \mathbf{W}_2^{(4)} \, \sigma\!\left( \mathbf{W}_1^{(4)} \boldsymbol{\phi} + \mathbf{b}_1^{(4)} \right) + \mathbf{b}_2^{(4)}
\end{align}
The encoded features are concatenated to form the composite feature vector serving as input to the temporal processor:
\begin{equation}
\mathbf{f} = \begin{bmatrix} \mathbf{f}_1 \\ \mathbf{f}_2 \\ \mathbf{f}_3 \\ \mathbf{f}_4 \end{bmatrix} \in \mathbb{R}^{d_f}
\end{equation}
where $d_f = \sum_{i=1}^{4} d_i$ denotes the total feature dimension, with $d_i$ being the output dimension of encoder $i$.

\subsection{Temporal Processing}
The temporal processor captures dynamics at multiple timescales through two parallel LSTM networks and a multi-head attention mechanism.

The first LSTM processes the complete feature sequence $\{\mathbf{f}_k\}_{k=t-T_h+1}^{t}$ to capture immediate dynamics.

At each timestep $k$, the LSTM updates its hidden state according to:

\begin{align}
\mathbf{i}_k^{(1)} &= \sigma_g\!\left( \mathbf{W}_{i}^{(1)} \mathbf{f}_k + \mathbf{U}_{i}^{(1)} \mathbf{h}_{k-1}^{(1)} + \mathbf{b}_{i}^{(1)} \right) \\[3pt]
\mathbf{g}_k^{(1)} &= \sigma_g\!\left( \mathbf{W}_{g}^{(1)} \mathbf{f}_k + \mathbf{U}_{g}^{(1)} \mathbf{h}_{k-1}^{(1)} + \mathbf{b}_{g}^{(1)} \right) \\[3pt]
\mathbf{o}_k^{(1)} &= \sigma_g\!\left( \mathbf{W}_{o}^{(1)} \mathbf{f}_k + \mathbf{U}_{o}^{(1)} \mathbf{h}_{k-1}^{(1)} + \mathbf{b}_{o}^{(1)} \right) \\[3pt]
\tilde{\mathbf{c}}_k^{(1)} &= \tanh\!\left( \mathbf{W}_{c}^{(1)} \mathbf{f}_k + \mathbf{U}_{c}^{(1)} \mathbf{h}_{k-1}^{(1)} + \mathbf{b}_{c}^{(1)} \right) \\[3pt]
\mathbf{c}_k^{(1)} &= \mathbf{g}_k^{(1)} \odot \mathbf{c}_{k-1}^{(1)} + \mathbf{i}_k^{(1)} \odot \tilde{\mathbf{c}}_k^{(1)} \\[3pt]
\mathbf{h}_k^{(1)} &= \mathbf{o}_k^{(1)} \odot \tanh\!\left( \mathbf{c}_k^{(1)} \right)
\end{align}

where $\sigma_g(\cdot)$ denotes the sigmoid function, $\mathbf{i}_k^{(1)}$, $\mathbf{g}_k^{(1)}$, and $\mathbf{o}_k^{(1)}$ represent the input, forget, and output gates respectively, $\mathbf{c}_k^{(1)}$ is the cell state, and $\mathbf{h}_k^{(1)} \in \mathbb{R}^{d_h}$ is the hidden state.

The second LSTM operates on a downsampled sequence with stride $\tau$ to capture slower dynamics and longer-term patterns. Let $\mathcal{T}_\tau = \{t - T_h + 1, t - T_h + 1 + \tau, \ldots, t\}$ denote the downsampled time indices. The second LSTM processes $\{\mathbf{f}_k\}_{k \in \mathcal{T}_\tau}$ using the same gating mechanism with independent parameters $\{\mathbf{W}^{(2)}, \mathbf{U}^{(2)}, \mathbf{b}^{(2)}\}$, yielding hidden state $\mathbf{h}_t^{(2)} \in \mathbb{R}^{d_h}$.

The final hidden states from both LSTMs are concatenated to form the composite temporal feature vector:
\begin{equation}
\mathbf{z} = \begin{bmatrix} \mathbf{h}_t^{(1)} \\ \mathbf{h}_t^{(2)} \end{bmatrix} \in \mathbb{R}^{2d_h}
\end{equation}

The concatenated features are further processed through a multi-head attention mechanism. We employ $N_h = 3$ attention heads, where each head computes scaled dot-product attention over the temporal feature vector $\mathbf{z}$.

For each head $h \in \{1, \ldots, N_h\}$, the query, key, and value projections are computed as:
\begin{align}
\mathbf{Q}^{(h)} &= \mathbf{W}_Q^{(h)} \mathbf{z}, \quad \mathbf{K}^{(h)} = \mathbf{W}_K^{(h)} \mathbf{z}, \quad \mathbf{V}^{(h)} = \mathbf{W}_V^{(h)} \mathbf{z}
\end{align}
where $\mathbf{W}_Q^{(h)}, \mathbf{W}_K^{(h)}, \mathbf{W}_V^{(h)} \in \mathbb{R}^{d_k \times 2d_h}$ are learnable projection matrices and $d_k = 2d_h / N_h$ is the dimension per head.

The attention output for each head is:
\begin{equation}
\mathbf{a}^{(h)} = \text{softmax}\left( \frac{\mathbf{Q}^{(h)} {\mathbf{K}^{(h)}}^\top}{\sqrt{d_k}} \right) \mathbf{V}^{(h)}
\end{equation}

The outputs from all heads are concatenated and projected to form the final attended representation:
\begin{equation}
\mathbf{z}_{\text{att}} = \mathbf{W}_O \begin{bmatrix} \mathbf{a}^{(1)} \\ \vdots \\ \mathbf{a}^{(N_h)} \end{bmatrix} + \mathbf{z}
\end{equation}
where $\mathbf{W}_O \in \mathbb{R}^{2d_h \times 2d_h}$ is the output projection matrix. The attended feature vector $\mathbf{z}_{\text{att}} \in \mathbb{R}^{2d_h}$ serves as input to the mixture of experts control stage.

\subsection{Mixture of Experts Control}
\vspace{-10pt}
The final stage employs $M$ expert networks, each specialising in different control strategies. Each expert $i$ is a three-layer feedforward network that maps the attended temporal feature vector $\mathbf{z}_{\text{att}}$ to a multi-step control sequence of horizon $H$: 
\begin{equation}
\mathbf{u_p}^{(i)} = \mathbf{W}_3^{(e_i)} \, \sigma\!\left( \mathbf{W}_2^{(e_i)} \, \sigma\!\left( \mathbf{W}_1^{(e_i)} \mathbf{z}_{\text{att}} + \mathbf{b}_1^{(e_i)} \right) + \mathbf{b}_2^{(e_i)} \right) + \mathbf{b}_3^{(e_i)}
\end{equation}

where $\mathbf{u_p}^{(i)} \in \mathbb{R}^{H}$ is the control sequence predicted by expert $i$, $\sigma(\cdot)$ denotes the ReLU activation function, and $\{\mathbf{W}_j^{(e_i)}, \mathbf{b}_j^{(e_i)}\}$ are learnable parameters for expert $i$ and layer $j$. Layer normalisation is applied after each hidden layer.

A gating network computes soft weights for expert combination based on the attended temporal features:
\begin{align}
\boldsymbol{\gamma} &=\mathbf{W}_g \mathbf{z}_{\text{att}} + \mathbf{b}_g \\[3pt]
\alpha_i &= \frac{\exp(\gamma_i)}{\sum_{j=1}^{M} \exp(\gamma_j)}, \quad i \in \{1, \ldots, M\}
\end{align}
where $\boldsymbol{\gamma} \in \mathbb{R}^{M}$ are the gate logits and $\alpha_i$ are the resulting mixture weights satisfying $\sum_{i=1}^{M} \alpha_i = 1$.

The final control sequence is computed as the weighted combination of expert outputs:
\begin{equation}
\hat{\mathbf{u_p}} = \sum_{i=1}^{M} \alpha_i \mathbf{u_p}^{(i)} \in \mathbb{R}^{H}
\end{equation}

The system evolves according to its dynamics in Eq.~\eqref{eq:state_dynamics}, producing new state measurements that are fed back to the controller, and the process repeats.

\section{Training Framework}
\vspace{-10pt}
We train the Generalist Controller on expert demonstrations from 25 dynamical systems spanning orders $n \in \{2, 3, 4\}$.

\subsection{Training Data Generation}
\label{sec:data_generation}
\vspace{-10pt}
Training data is generated by simulating each system under LQI control with step reference signals. Step amplitudes are sampled from $[-20, 20]$ using non-uniform discrete intervals, coarse spacing (intervals of 5 or 10) in some regions and finer resolution (intervals of 0.2 or 1) in others to ensure coverage across both small and large reference changes. For each system-amplitude pair, trajectories are recorded over a fixed simulation horizon, yielding state, reference, and control tuples. This procedure produces approximately 314,630 training demonstrations across all 25 systems.

\subsection{Training Configuration}
\vspace{-10pt}
Training sequences employ sliding windows with history $H = 8$ and prediction horizon $K = 3$. States $\mathbf{x}_{t-H:t} \in \mathbb{R}^{H \times n}$ are zero-padded to maximum dimension $n_{\max} = 4$, with validity mask $\mathbf{m} \in \{0,1\}^{H \times n_{\max}}$ indicating real versus padded dimensions.  The history length $H=8$ was selected as a trade-off between capturing sufficient temporal information to infer system dynamics and maintaining low computational cost. Similarly, the prediction horizon $K=3$ provides short-term foresight suitable for receding-horizon control while avoiding error accumulation associated with longer open-loop predictions. These hyperparameters were fixed across all systems to ensure consistency and to avoid system-specific tuning. No claim is made that these values are optimal, rather, they represent pragmatic choices that were found to work reliably across the considered benchmark systems.

The hierarchical architecture comprises: (i) encoders (state encoders) with masked attention processing state trajectories, (ii) parameter and reference encoders mapping to a common latent space, (iii) parallel multi-scale LSTMs capturing temporal dependencies at different resolutions, (iv) multi-head attention for temporal aggregation, and (v) a mixture-of-experts controller with $M = 3$ experts:
\begin{equation}
\mathbf{u}_{t:t+K} = \sum_{i=1}^{M} \alpha_i(\mathbf{h}_t) \cdot \mathcal{C}_i(\mathbf{h}_t)
\end{equation}
where $\alpha_i$ are learned gating weights and $\mathcal{C}_i$ represents expert $i$.

Training minimises mean squared error with AdamW optimiser ($\eta = 10^{-3}$, cosine annealing), gradient clipping ($\|\nabla\|_2 \leq 1.0$), and weight decay ($\lambda = 10^{-5}$). The model contains approximately 100,000 parameters and trains in six hours on a single GPU.

\section{Simulation and Experimental Results}
\vspace{-10pt}
This section evaluates the proposed Generalist Controller across diverse dynamical systems. We first present the benchmark systems, then report simulation results on six representative systems spanning different orders and dynamic behaviours. We then validate the approach through hardware experiments on a Crazyflie 2.1+ nano-quadrotor, demonstrating successful sim-to-real transfer. Finally, we assess the controller's robustness under perturbed conditions not encountered during training, including actuator saturation, disturbances, and measurement noise.

\subsection{Benchmark Systems}
\vspace{-10pt}
The 25 benchmark systems encompass diverse characteristics and applications, fundamental mechanical systems (Mass-Spring-Damper, Damped Oscillator, Two-Mass Spring), electrical circuits (RLC Circuit), aerospace systems (Crazyflie quadrotor, Aircraft Longitudinal dynamics), robotic platforms (Differential Drive, Servo System), process control systems (Two-Tank System, Coupled Tanks, HVAC Zone), and systems with challenging dynamics (Non-Minimum Phase, Negative Stiffness, unstable integrators). Several of these systems were previously analysed in \cite{agyei2025deep}, including the CSTR Chemical Reactor, nonlinear AUV yaw dynamics, and Two-Mass Spring system, providing established baselines for comparison.

\subsection{Simulation Results}
\vspace{-10pt}
In the simulation section, we further assess and present results for six representative systems selected from the benchmark systems: the CSTR Chemical Process (non-minimum phase), an Unstable System, Boeing 747 Aircraft (longitudinal dynamics), Hydraulic Actuator, Nonlinear AUV (REMUS yaw dynamics), and Two-Mass Spring. Table~\ref{tab:benchmark_systems} presents these results. In the first two columns, system descriptions and state-space equations are provided. The third column shows controller performance on step response tracking with unit step references, comparing the proposed Generalist Controller against the LQI method. Furthermore, in the fourth column, we present simulation results for sinusoidal signal tracking from non-zero initial states not encountered during training, to assess the generalisability and resilience of the algorithm when initial conditions change. The controller demonstrates similar successful performance on ramp following and composite reference signals, with none of these complex references encountered by the model during training. Due to space constraints, the paper presents representative step-response and sinusoidal tracking results, while additional evaluations under alternative reference trajectories were conducted and yielded consistent qualitative behaviour.

Table~\ref{tab:benchmark_performance} reports a direct quantitative comparison between the proposed Generalist Controller and the system-specific LQI controllers using standard control performance metrics    \cite{sarhadi2025standardcriteria}. Across all simulated benchmark systems, the proposed Generalist Controller (GC) achieves performance comparable to LQI in terms of rise time, overshoot, and steady-state error, despite being deployed without per-system tuning. In several cases, including the hydraulic actuator and Boeing~747 longitudinal dynamics, the GC exhibits similar or slightly improved settling times relative to LQI, while maintaining comparable control effort. For more challenging systems, such as the non-minimum-phase and two-mass spring benchmarks, the GC demonstrates slightly increased settling times. Overall, the results indicate that the GC preserves the essential closed-loop behaviour of classical LQI controllers while providing a single deployable policy across heterogeneous systems.

\begin{table}[!t]
\caption{Performance metrics comparison between LQI and the Proposed Method}
\label{tab:benchmark_performance}
\centering
\scriptsize
\setlength{\tabcolsep}{2.3pt}
\renewcommand{\arraystretch}{1.05}
\begin{tabular}{lccccccc}
\bottomrule
Metric & Unst. & NMP & H.Act & Yaw & B-747 & TMS & Crazyflie\\
\midrule
$t_r$ (LQI) & 1.75 & 1.05 & 4.35 & 2.50 & 1.20 & 1.70 & 2.15 \\
$t_r$ (GC)  & 1.95 & 1.05 & 3.95 & 2.45 & 1.40 & 1.75 & 1.85 \\

$M_p$ (LQI) & 5.5  & 0.4  & 0.0  & 0.8  & 2.3  & 11.5 & 1.60 \\
$M_p$ (GC)  & 5.4  & 0.7  & 0.2  & 0.4  & 2.0  & 11.1 & 2.00 \\

$t_s$ (LQI) & 5.05 & 2.20 & 8.15 & 4.25 & 2.85 & 7.25 & 4.05 \\
$t_s$ (GC)  & 6.35 & 2.25 & 7.80 & 5.10 & 2.35 & 9.50 & 5.55 \\

$e_{\text{ss}}$ (LQI) & 0.00 & 0.00 & 0.00 & 0.00 & 0.00 & 0.00 & 0.00 \\
$e_{\text{ss}}$ (GC)  & 0.00 & 0.00 & 0.00 & 0.00 & 0.00 & 0.00 & 0.00 \\

ISE (LQI) & 1.21 & 1.09 & 1.40 & 1.14 & 0.82 & 1.88 & 0.87 \\
ISE (GC)  & 1.28 & 1.12 & 1.53 & 1.17 & 0.92 & 1.99 & 0.79 \\

ITAE (LQI) & 1.8 & 0.8 & 4.8 & 2.2 & 0.7 & 3.9 & 1.4 \\
ITAE (GC)  & 2.9 & 1.0 & 5.0 & 4.2 & 1.1 & 5.1 & 2.50 \\

$u_{\max}$ (LQI) & 0.73 & 1.97 & 0.14 & 1.10 & 1.74 & 1.36 & 0.08 \\
$u_{\max}$ (GC)  & 0.65 & 1.96 & 0.13 & 1.12 & 1.58 & 1.34 & 0.10 \\
\bottomrule
\end{tabular}
\end{table}

\subsection{Experimental Results}
\begin{figure}[b]
\centering
\includegraphics[width=0.9\linewidth]{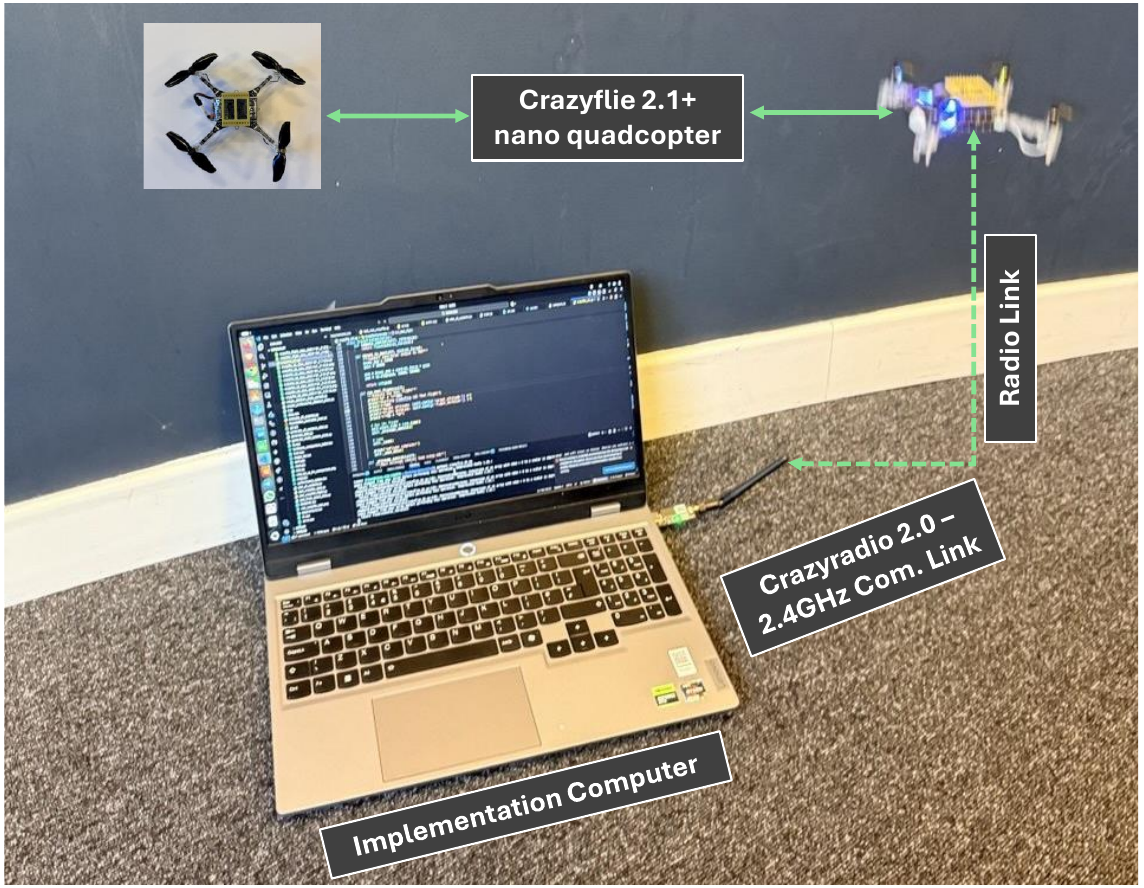}
\caption{Hardware validation: Crazyflie in flight with the proposed algorithm}
\label{fig:hardware_setup}
\end{figure}

\begin{figure}[!t]
\centering
\begin{minipage}[b]{0.48\linewidth}
\centering
\includegraphics[width=\linewidth]{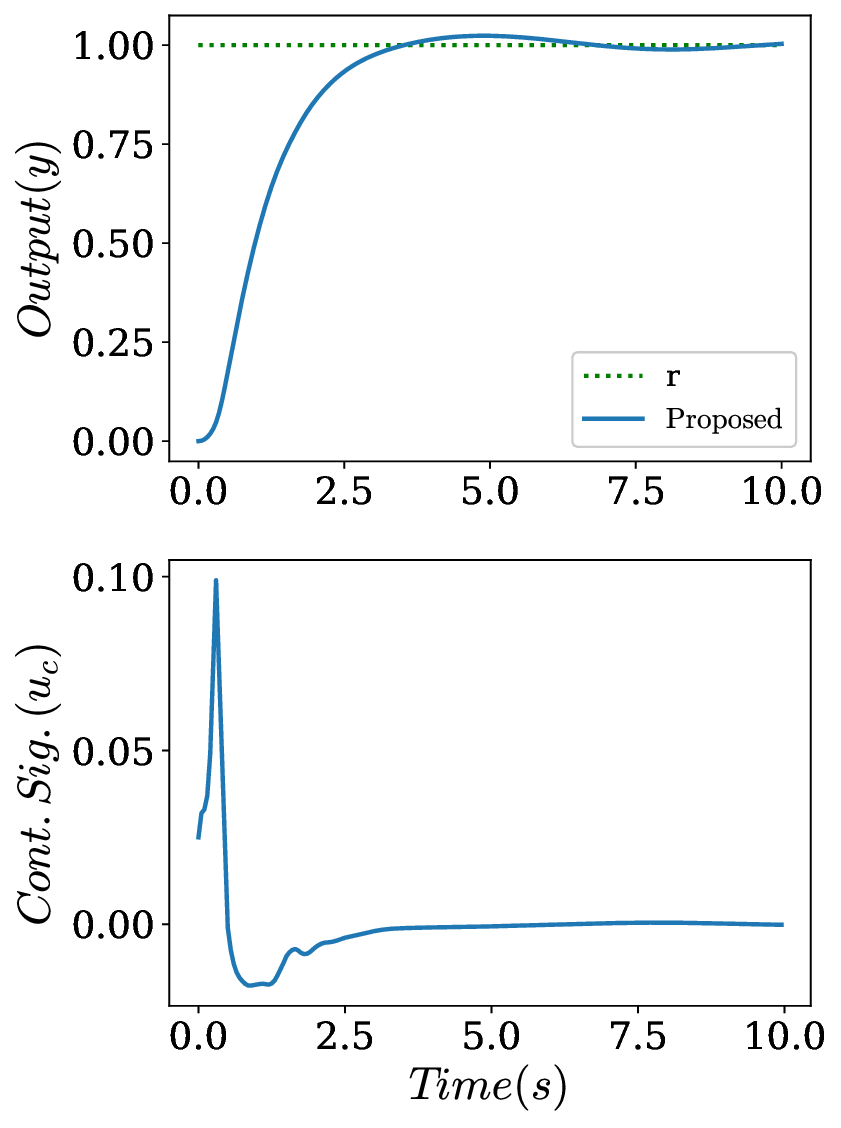}

\end{minipage}
\hfill
\begin{minipage}[b]{0.48\linewidth}
\centering
\includegraphics[width=\linewidth]{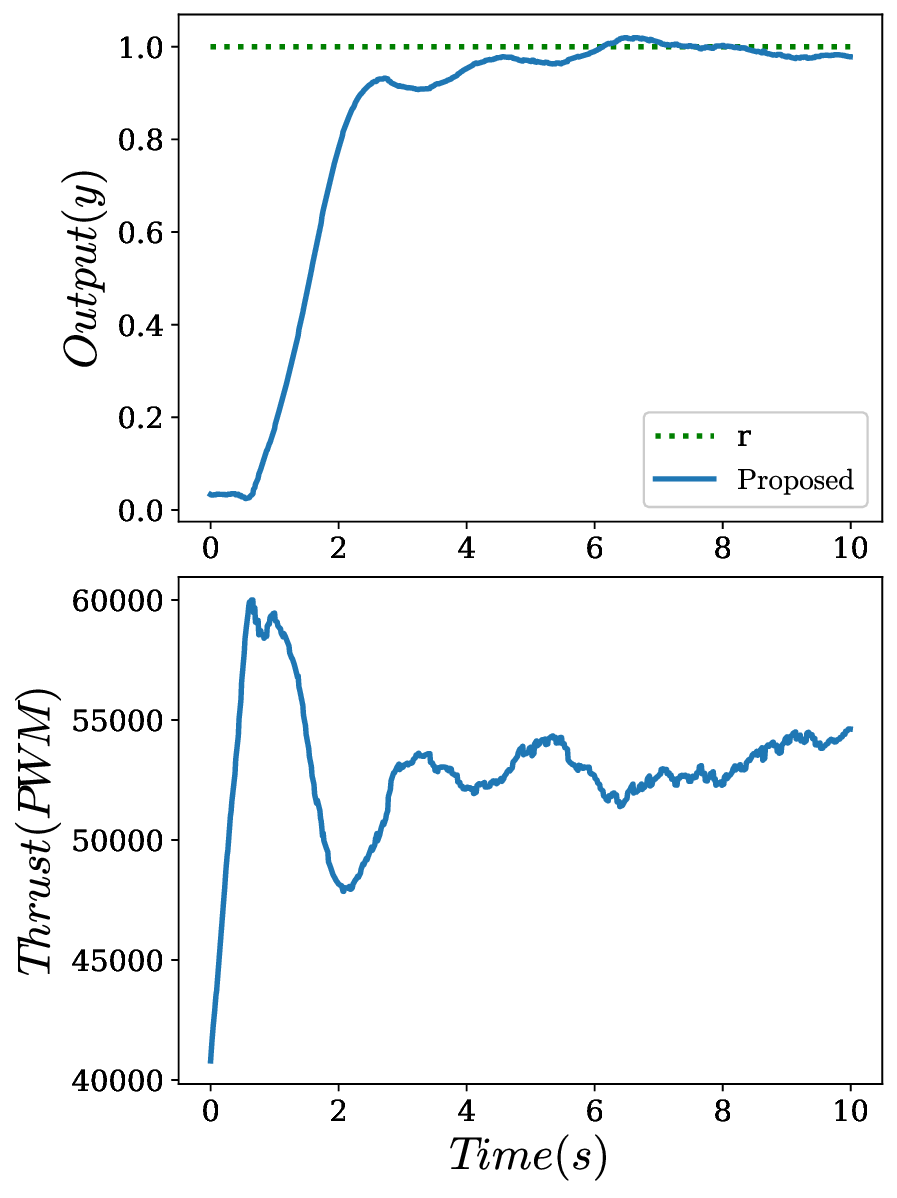}

\end{minipage}
\caption{Sim-to-real transfer: Crazyflie 2.1+ altitude control in (Left) simulation and (Right) hardware.}
\label{fig:crazyflie_simtoreal}
\end{figure}

\vspace{-10pt}
We validate the proposed controller on hardware using a Crazyflie 2.1+ nano-quadrotor. Figure~\ref{fig:hardware_setup} shows the experimental setup. The Generalist Controller runs on a Linux computer with Python implementation, communicating with the quadcopter via the Crazyradio 2.0 USB dongle operating at 2.4GHz. The controller receives state feedback at 100Hz and computes control commands in real-time, demonstrating successful deployment of the learned policy on resource-constrained embedded systems.
Figure~\ref{fig:crazyflie_simtoreal} presents a direct comparison between simulation and hardware performance for altitude control. The physical system tracks the reference with similar transient behaviour and steady-state accuracy to simulation, confirming that the controller generalises effectively to real hardware without fine-tuning, despite unmodelled aerodynamic effects, sensor noise, and actuator delays.

\subsection{Further Evaluation Under Perturbations}
\label{subsec:Further_Evaluation}
\vspace{-10pt}
To assess the robustness and generalisation capability of the proposed Generalist Controller beyond nominal operating conditions, we conduct a series of experiments under perturbed test scenarios not encountered during training.

\begin{figure}[!t]
\centering
\begin{minipage}[b]{0.48\linewidth}
\centering
\includegraphics[width=\linewidth]{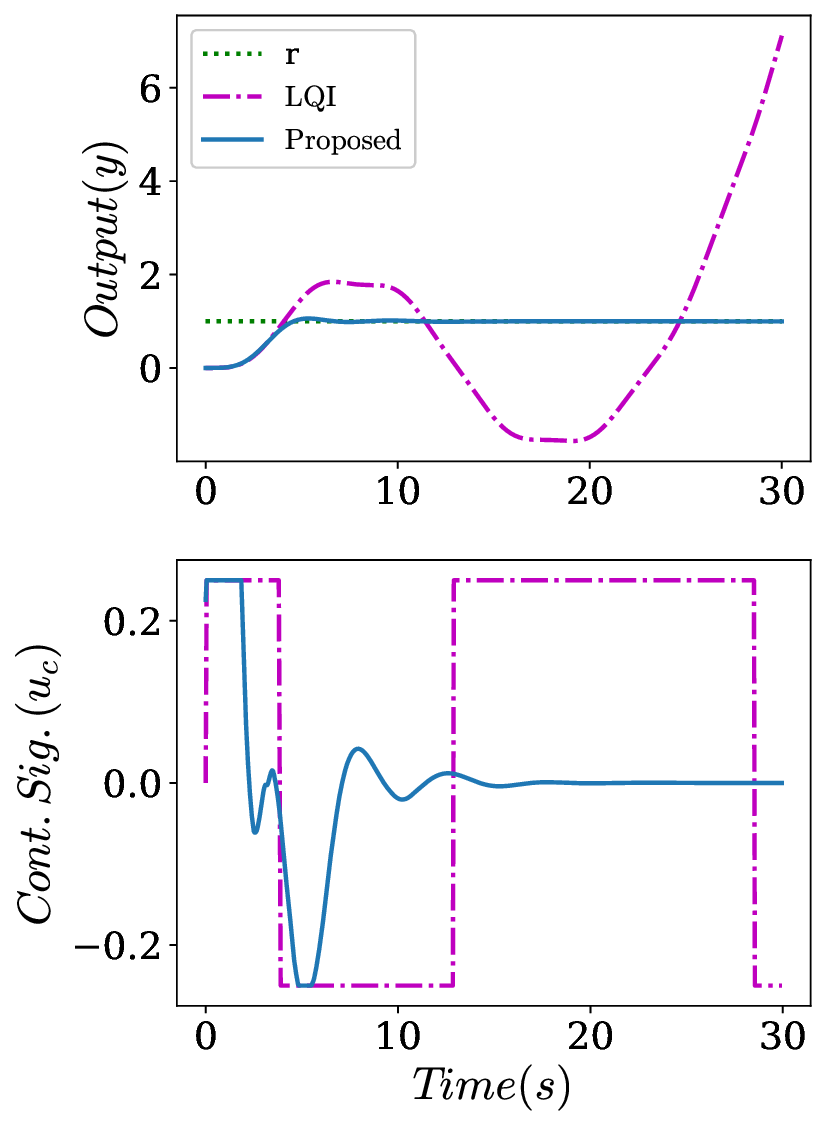}

\end{minipage}
\hfill
\begin{minipage}[b]{0.48\linewidth}
\centering
\includegraphics[width=\linewidth]{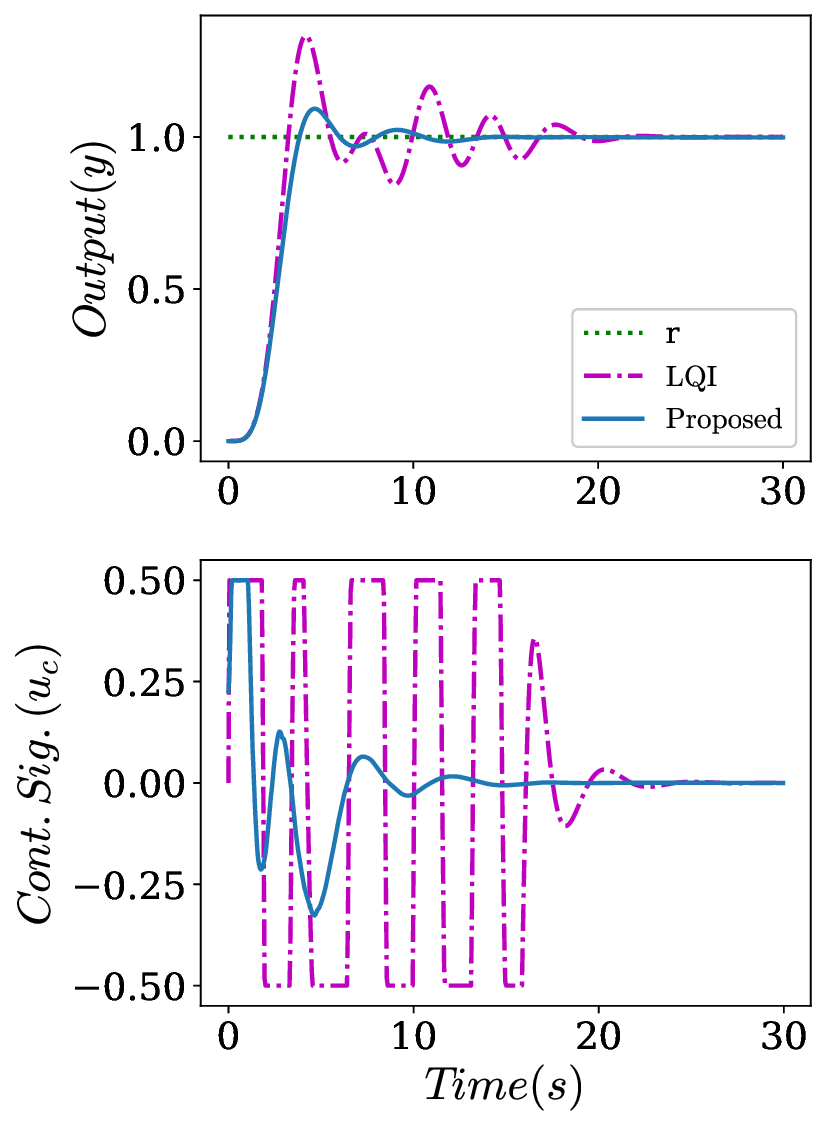}

\end{minipage}
\caption{TMS system under actuator amplitude saturation. Left: $|u_c| \leq 0.25$, Right: $|u_c| \leq 0.5$}
\label{fig:tmssat}
\end{figure}

We first evaluate the controller under actuator amplitude saturation constraints applied to the challenging Two-Mass Spring (TMS) system. Fig.~\ref{fig:tmssat} presents results for two saturation limits: $|u_c| \leq 0.5$ and $|u_c| \leq 0.25$. Under moderate saturation ($|u_c| \leq 0.5$), the proposed method successfully tracks the reference with minimal overshoot, whilst the LQI controller exhibits significant oscillatory behaviour before eventually converging. When the saturation constraint is tightened to $|u_c| \leq 0.25$, the LQI controller becomes unstable and diverges, whereas the proposed method maintains stable tracking despite the limited control authority. These results demonstrate that the baseline LQI controller fails to maintain performance under actuator constraints, whereas the learned policy sustains stable tracking despite these unseen limitations.

\begin{figure}[b]
\centering
\includegraphics[width=0.55\linewidth]{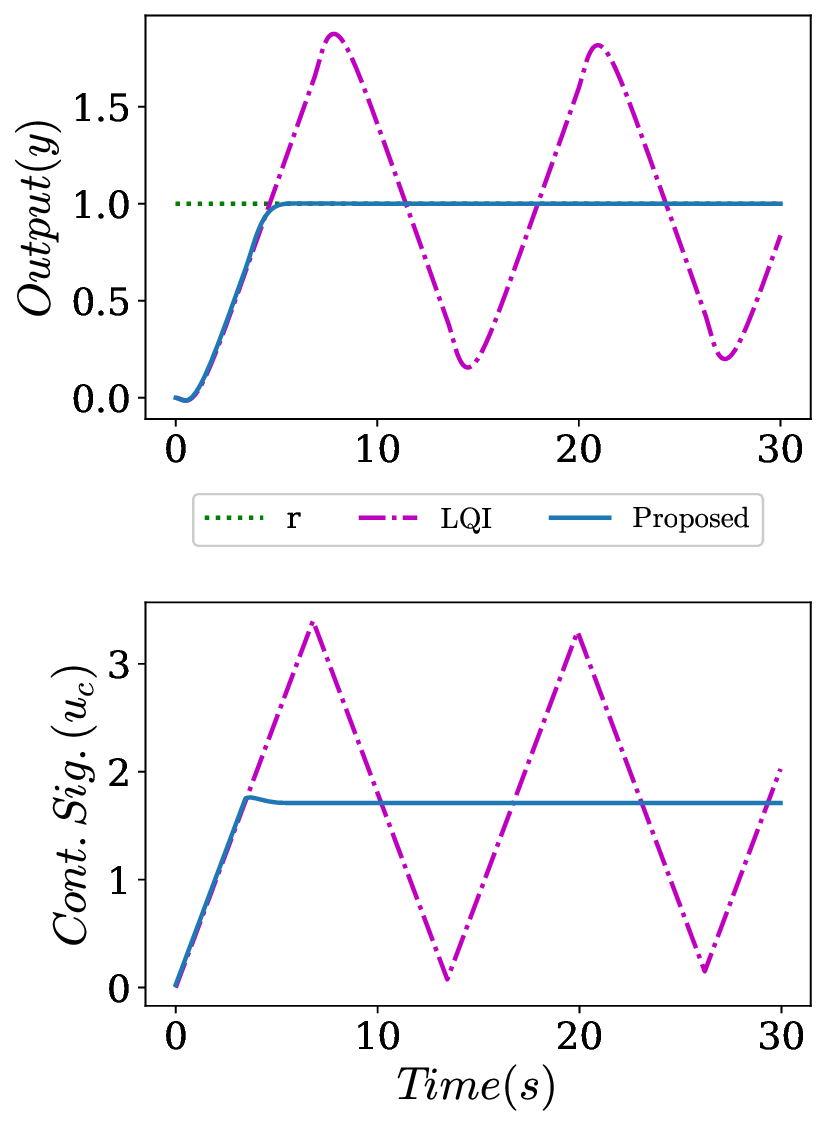}
\caption{CSTR Chemical process system under input rate saturation ($|\dot{u_c}| = 0.5$)}
\label{fig:nmprsat}
\end{figure}

We further evaluate performance under input rate saturation applied to the CSTR chemical process system. Figure~\ref{fig:nmprsat} shows the response when the rate of change of the control signal is constrained. The LQI controller, unable to adapt to the rate limitation, produces large oscillations in both output and control signal as it attempts corrections that are subsequently clipped, encountering the windup phenomenon. In contrast, the proposed method, though never exposed to rate constraints during training, maintains smooth reference tracking with a well-behaved control signal.

\begin{figure}[!t]
\centering
\begin{minipage}[b]{0.48\linewidth}
\centering
\includegraphics[width=\linewidth]{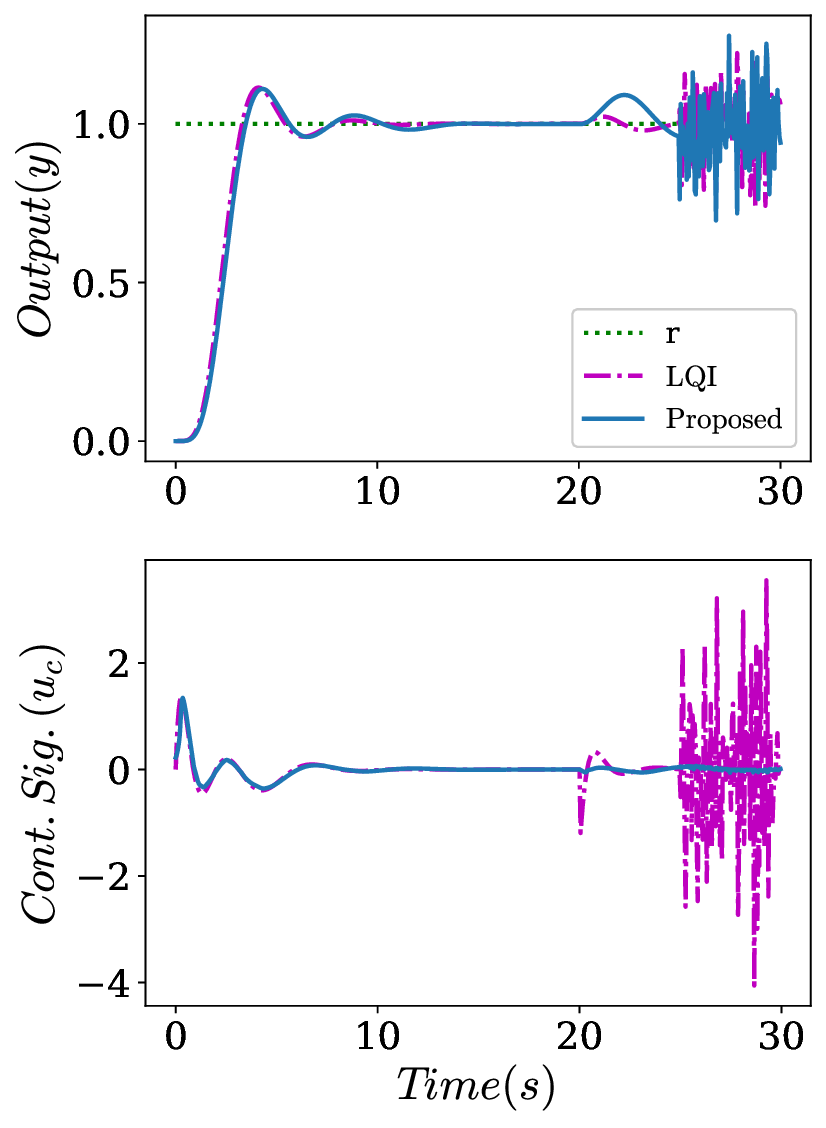}

\end{minipage}
\hfill
\begin{minipage}[b]{0.48\linewidth}
\centering
\includegraphics[width=\linewidth]{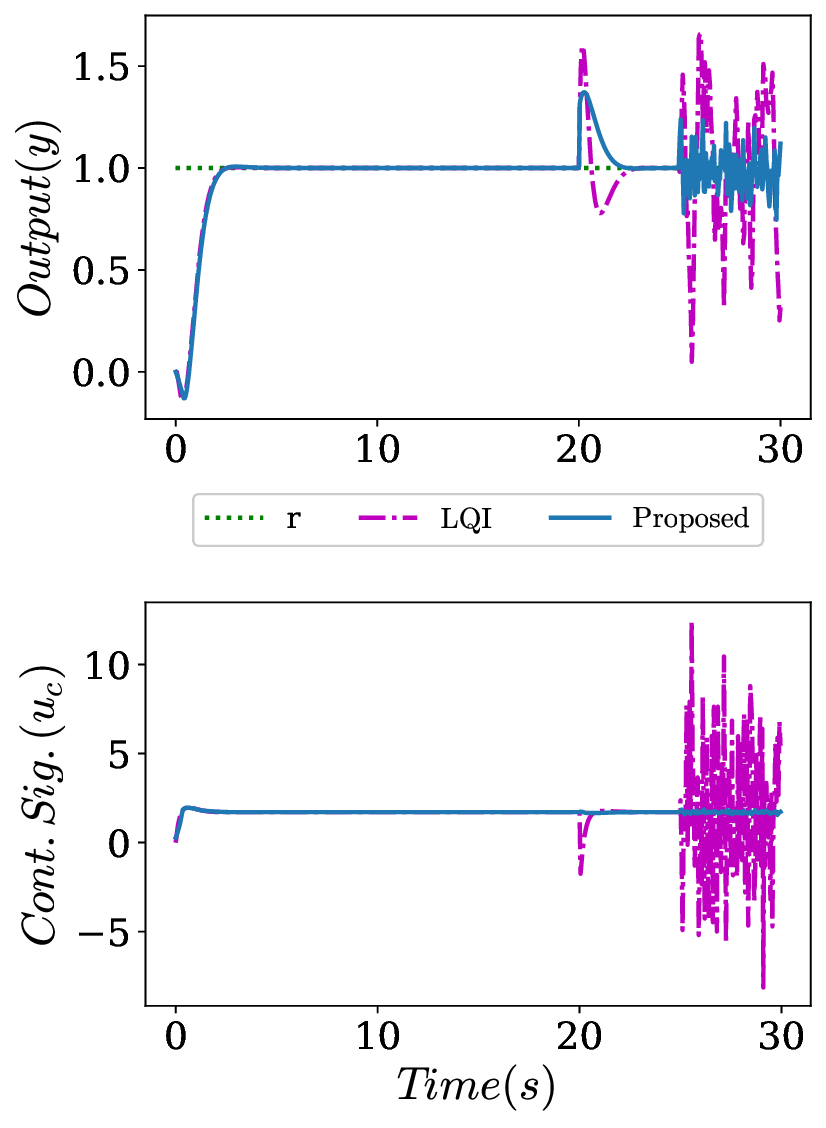}

\end{minipage}
\caption{Performance evaluation under combined measurement noise and external disturbance. Left: TMS. Right: CSTR}
\label{fig:noise}
\end{figure}

Finally, we assess performance under combined measurement noise and external disturbances for both TMS and CSTR systems. Fig.~\ref{fig:noise} presents the results with noise injected into state measurements after $t = 25$~s. For the CSTR system, the proposed method demonstrates superior performance compared to LQI, maintaining tighter tracking with reduced output variance despite the noisy measurements. The LQI controller exhibits highly aggressive control behaviour with large-amplitude oscillations in the control signal, which propagate to the output. For the TMS system, both controllers maintain stability under noise, though the proposed method shows slightly more variation in the output. However, examining the control signals reveals that the LQI controller demands significantly more aggressive actuation with peak-to-peak amplitudes substantially larger than those of the proposed method. This aggressive behaviour, whilst achieving marginally tighter tracking in the TMS case, would be problematic in practical applications where actuator wear, energy consumption, and excitation of unmodelled high-frequency dynamics are concerns.

These experiments demonstrate that the proposed controller generalises to operating conditions outside the training distribution, including actuator constraints and measurement corruption not seen during training. The successful performance under amplitude saturation, rate saturation, disturbance and noise injection provides evidence that the learned policy captures robust control principles rather than merely memorising input-output mappings from the training data.

\begin{table*}[!t]
\centering
\caption{\textsc{Results of the proposed generalist controller on multiple benchmarks.}}
\label{tab:benchmark_systems}
\setlength{\arrayrulewidth}{0.3pt}
\arrayrulecolor{gray!50}
\renewcommand{\arraystretch}{1.2}
\footnotesize
\begin{tabular}{|>{\centering\arraybackslash}m{2.2cm}|>{\centering\arraybackslash}m{6.1cm}|>{\centering\arraybackslash}m{3.7cm}|>
{\centering\arraybackslash}m{3.7cm}|}
\hline
\rowcolor{gray!10}
\textbf{System} & \textbf{State-Space Representation} & \textbf{Step Response} & \textbf{Sine Response} \\
\hline
\textbf{CSTR Chemical Process}\newline{\small\color{gray}(Non-Minimum Phase) \cite{agyei2025deep}} &
$\begin{array}{@{}l@{}}
\mathbf{\dot{x}_{p}} =
\begin{bmatrix}
0 & 1 \\
-5.47 & -4.719
\end{bmatrix} \mathbf{x_p} + \begin{bmatrix}
0 \\
1
\end{bmatrix} u_p \\[6pt]
y_p =
\begin{bmatrix}
3.199 & -1.135
\end{bmatrix} \mathbf{x_p}
\end{array}$ &
\includegraphics[width=0.75\linewidth]{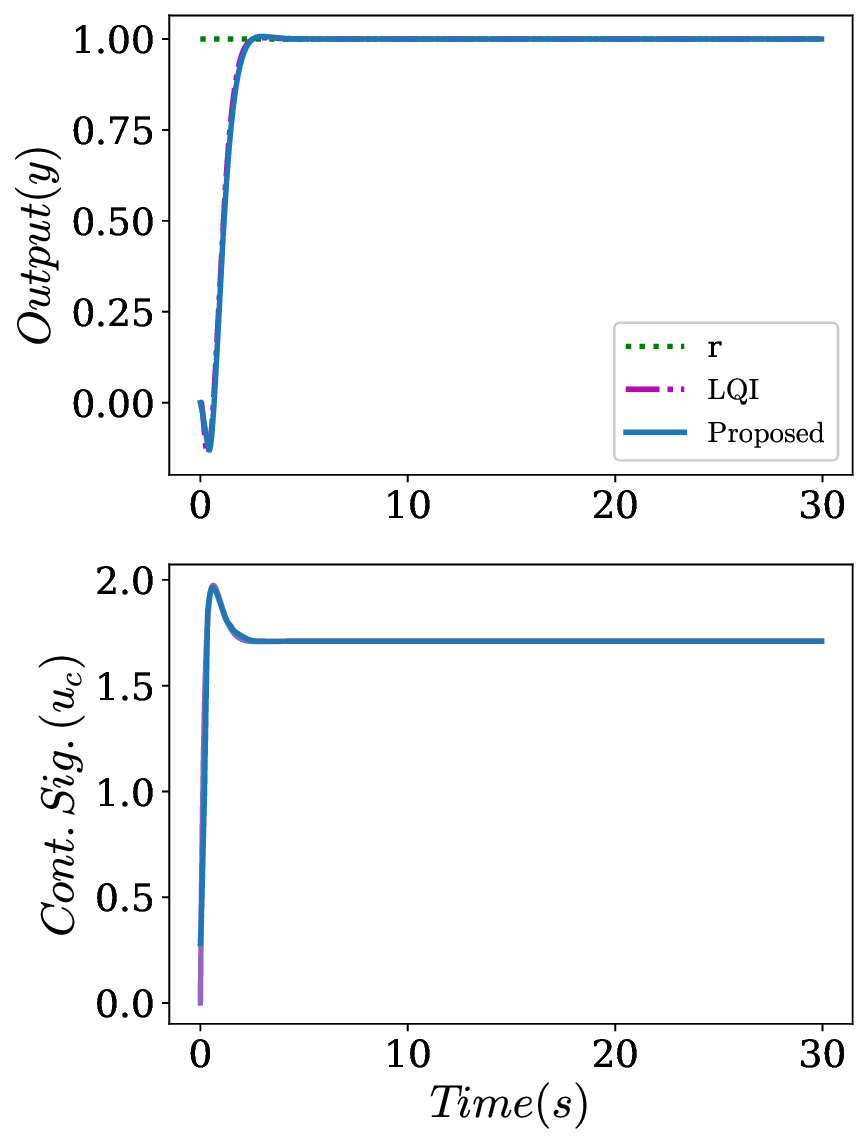} & \includegraphics[width=0.75\linewidth]{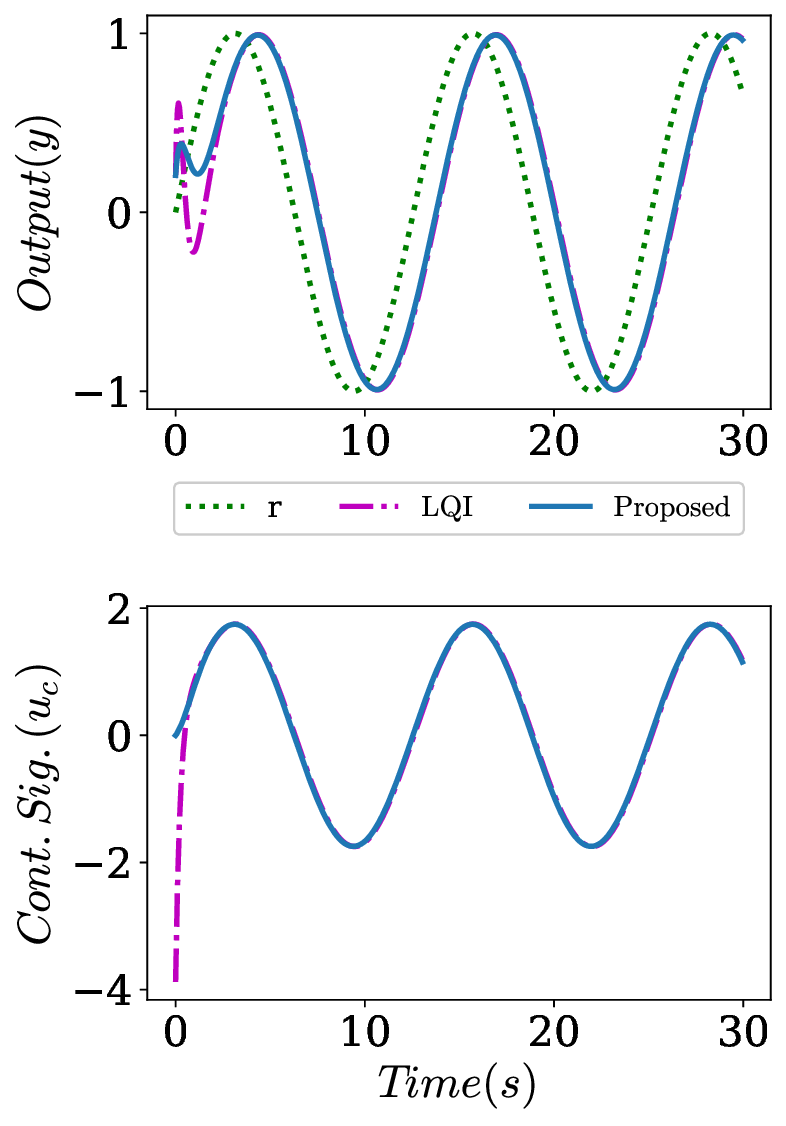} \\
\hline
\textbf{Unstable System}\newline{\small\color{gray}($G_p=\frac{1}{s(s-1)}$)} &
$\begin{array}{@{}l@{}}
\mathbf{\dot{x}_{p}} =
\begin{bmatrix}
0 & 1 \\
0 & 1
\end{bmatrix} \mathbf{x_p} +
\begin{bmatrix}
0 \\
1
\end{bmatrix} u_p \\[4pt]
y_p =
\begin{bmatrix}
1 & 0
\end{bmatrix} \mathbf{x_p}
\end{array}$ &
\includegraphics[width=0.75\linewidth]{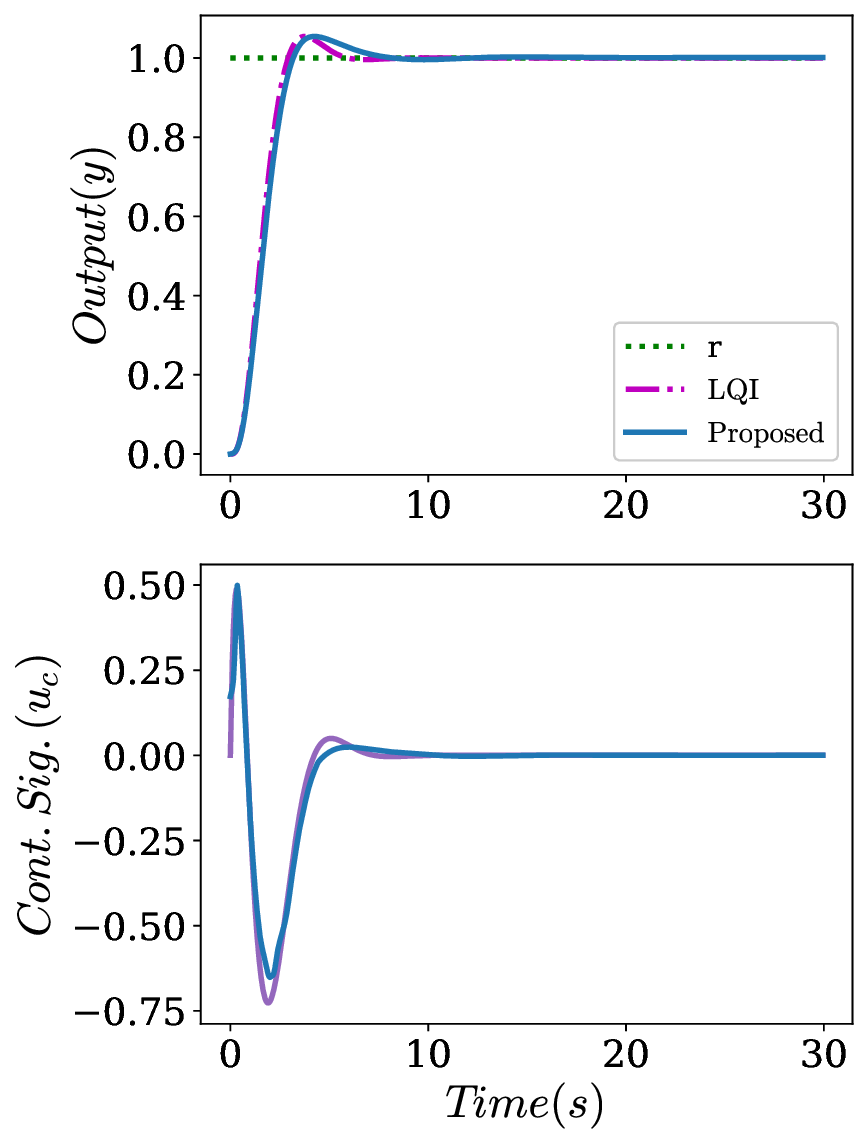} &
\includegraphics[width=0.75\linewidth]{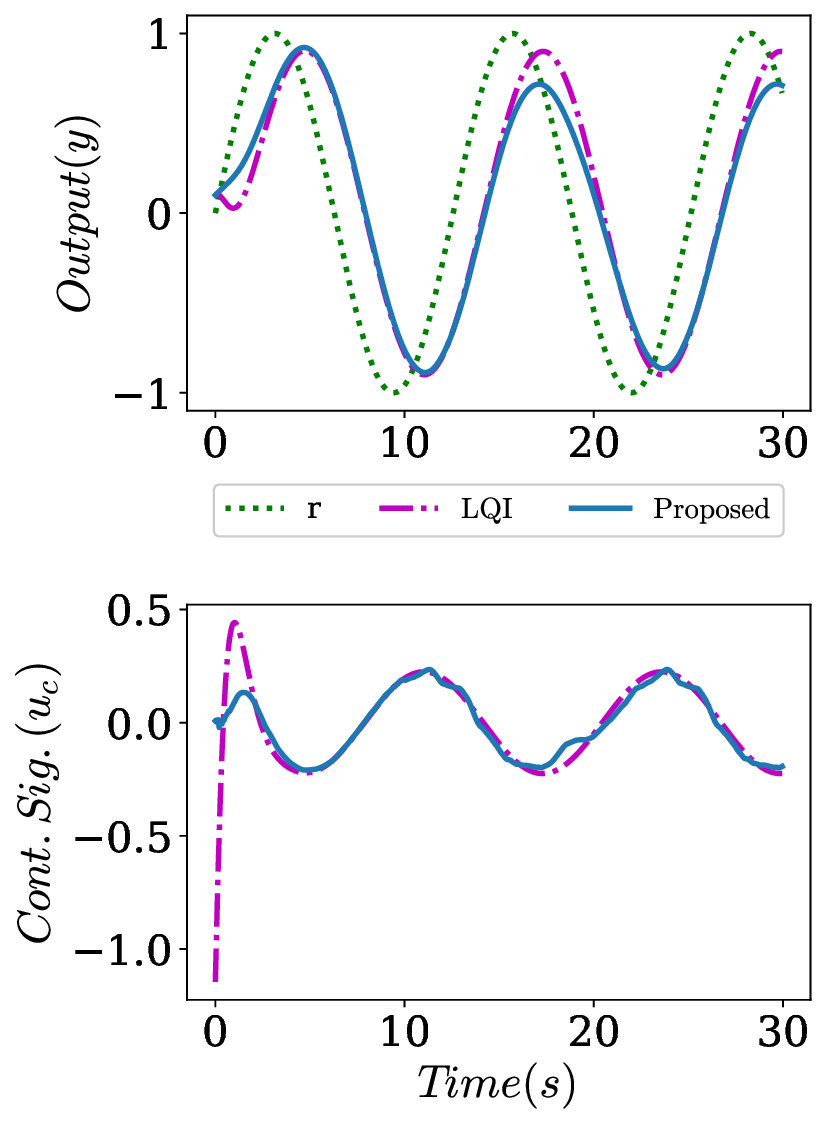}\\
\hline
\textbf{Boeing 747 Aircraft}\newline{\small\color{gray}(Longitudinal Dynamics) \cite{inanc2025neural}} &
$\begin{array}{@{}l@{}}
\mathbf{\dot{x}_{p}} =
\begin{bmatrix}
-0.32 & 0.86 \\
-0.93 & -0.43
\end{bmatrix} \mathbf{x_p} +
\begin{bmatrix}
-0.02 \\
-1.16
\end{bmatrix} u_p \\[4pt]
y_p =
\begin{bmatrix}
1 & 0
\end{bmatrix} \mathbf{x_p}
\end{array}$ &
\includegraphics[width=0.75\linewidth]{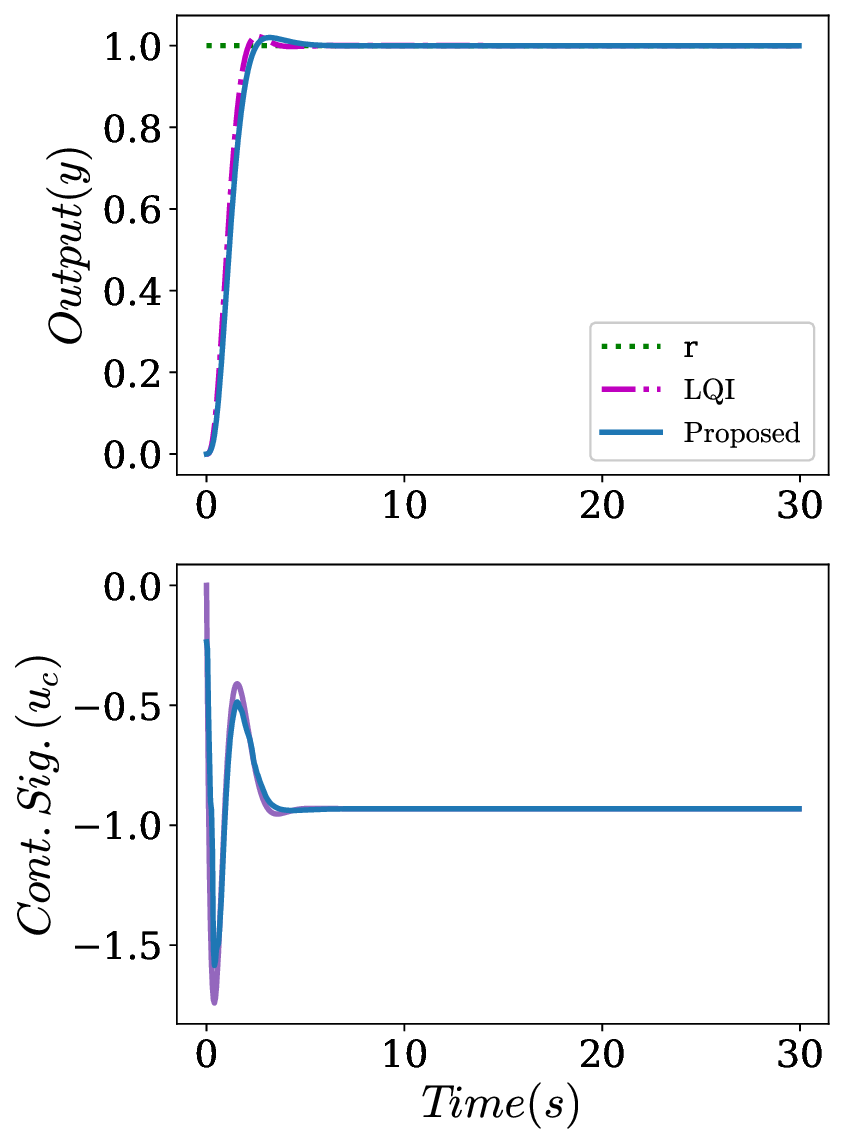} &
\includegraphics[width=0.75\linewidth]{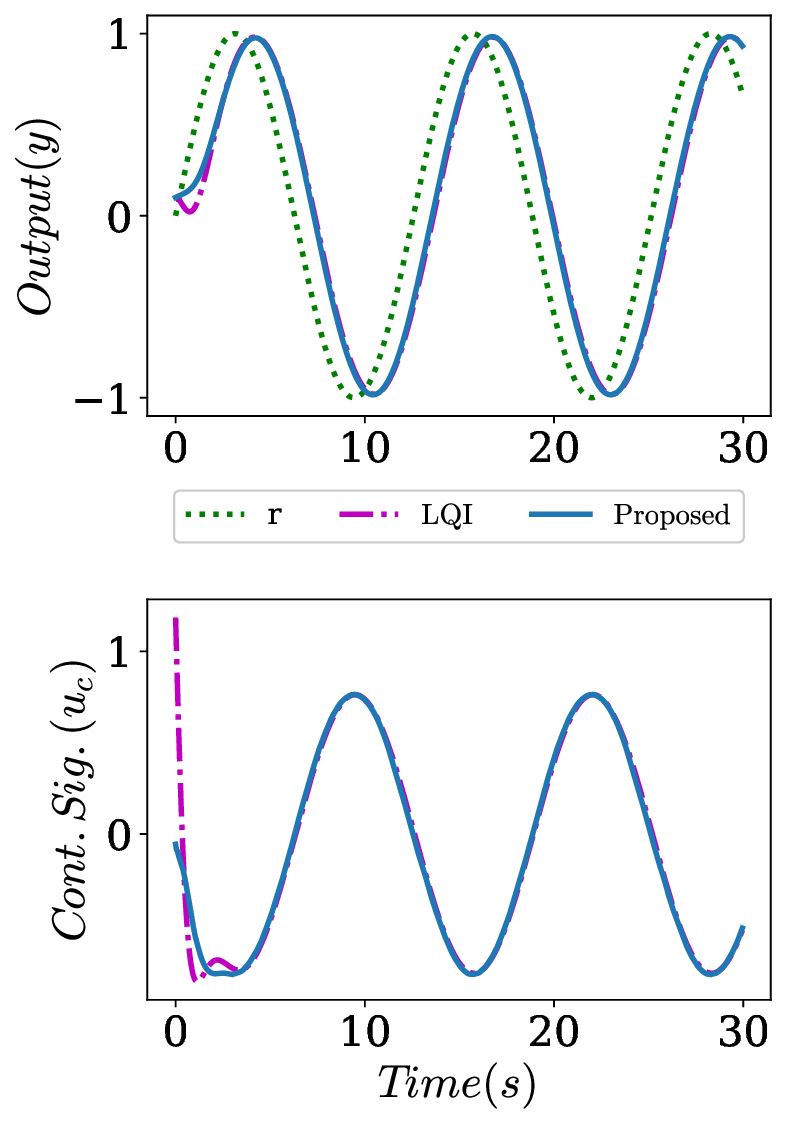}\\
\hline
\textbf{Hydraulic Actuator}\newline{\small\color{gray}(3D Servo Valve)} &
$\begin{array}{@{}l@{}}
\mathbf{\dot{x}_{p}} =
\begin{bmatrix}
0 & 1 & 0 \\
-10 & -1.167 & 25 \\
0 & 0 & -0.8
\end{bmatrix} \mathbf{x_p} +
\begin{bmatrix}
0 \\
0 \\
2.4
\end{bmatrix} u_p \\[4pt]
y_p =
\begin{bmatrix}
1 & 0 & 0
\end{bmatrix} \mathbf{x_p}
\end{array}$ &
\includegraphics[width=0.75\linewidth]{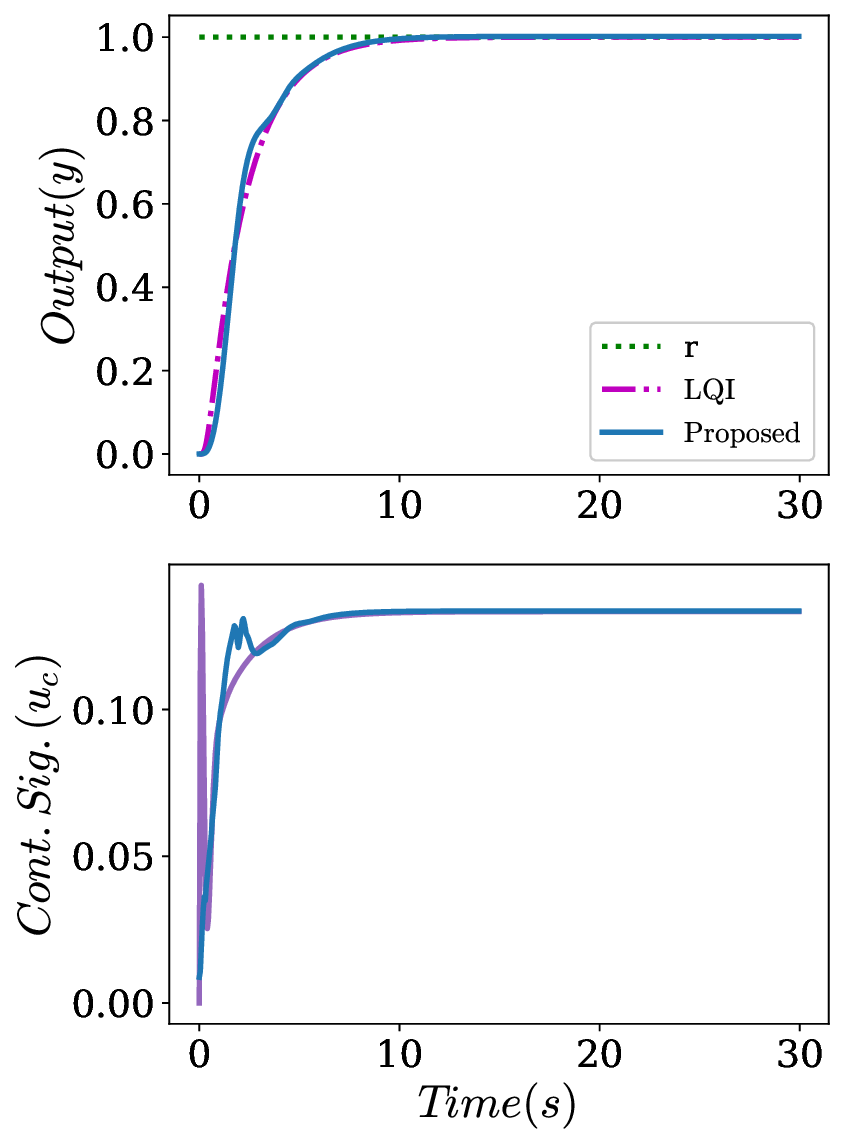} &
\includegraphics[width=0.75\linewidth]{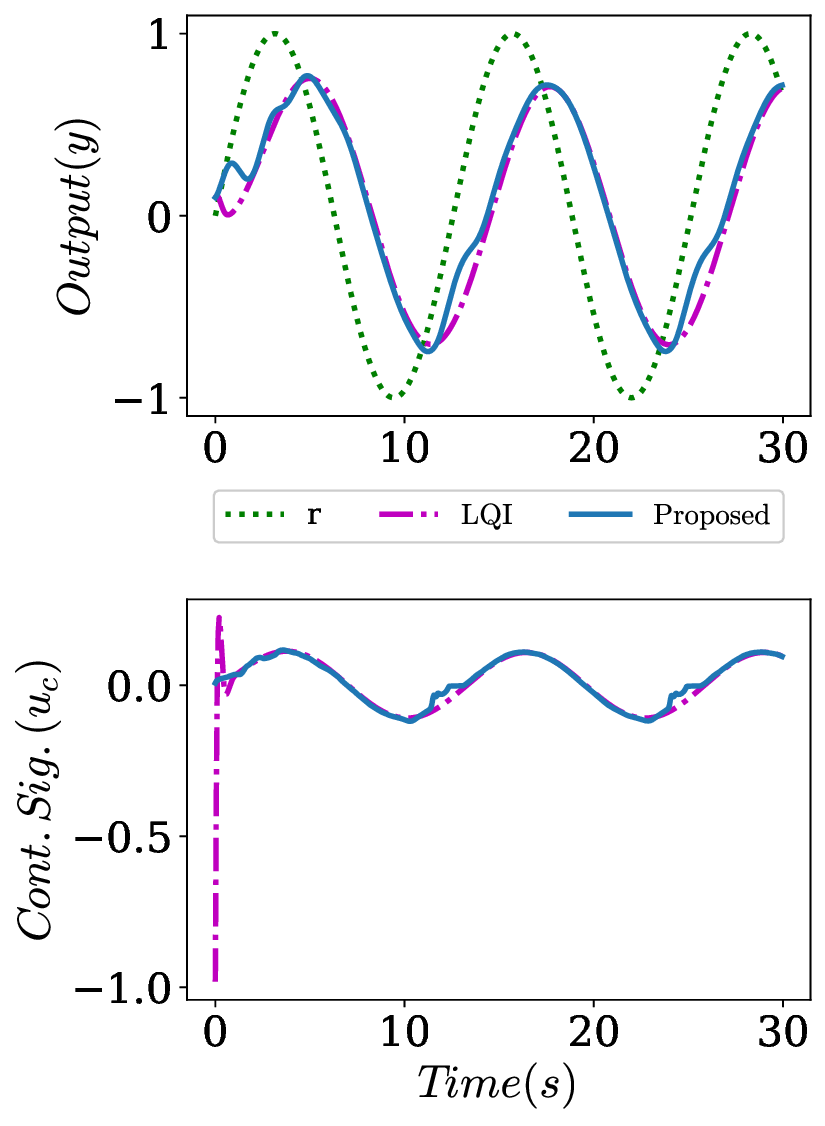}\\
\hline
\textbf{Nonlinear AUV}\newline{\small\color{gray}(REMUS Yaw Dynamics) \cite{agyei2025deep}} &
$\begin{array}{l}
(m - I_1)\dot{v} + (m\nu_g - I_4)\dot{r} = (Y_{uv}|u| + Y_{uuu}u^2)\nu\\
\qquad + (Y_{ur}|u| + Y_{uur}u^2)r - m\nu_0 u\dot{\psi}_0^2\\[0.05cm]
(m\nu_g - V_1)\dot{v} + (I_z - N_r)\dot{r} = (N_{uv}|u| + \\ N_{uuu}u^2)\nu + (N_{ur}|u| + N_{uur}u^2)r - m\nu_g u\dot{\psi}_0 + \\ N_{uur}\delta\delta_0^2\\[0.0cm]
\dot{\phi} = r
\end{array}$ &
\includegraphics[width=0.75\linewidth]{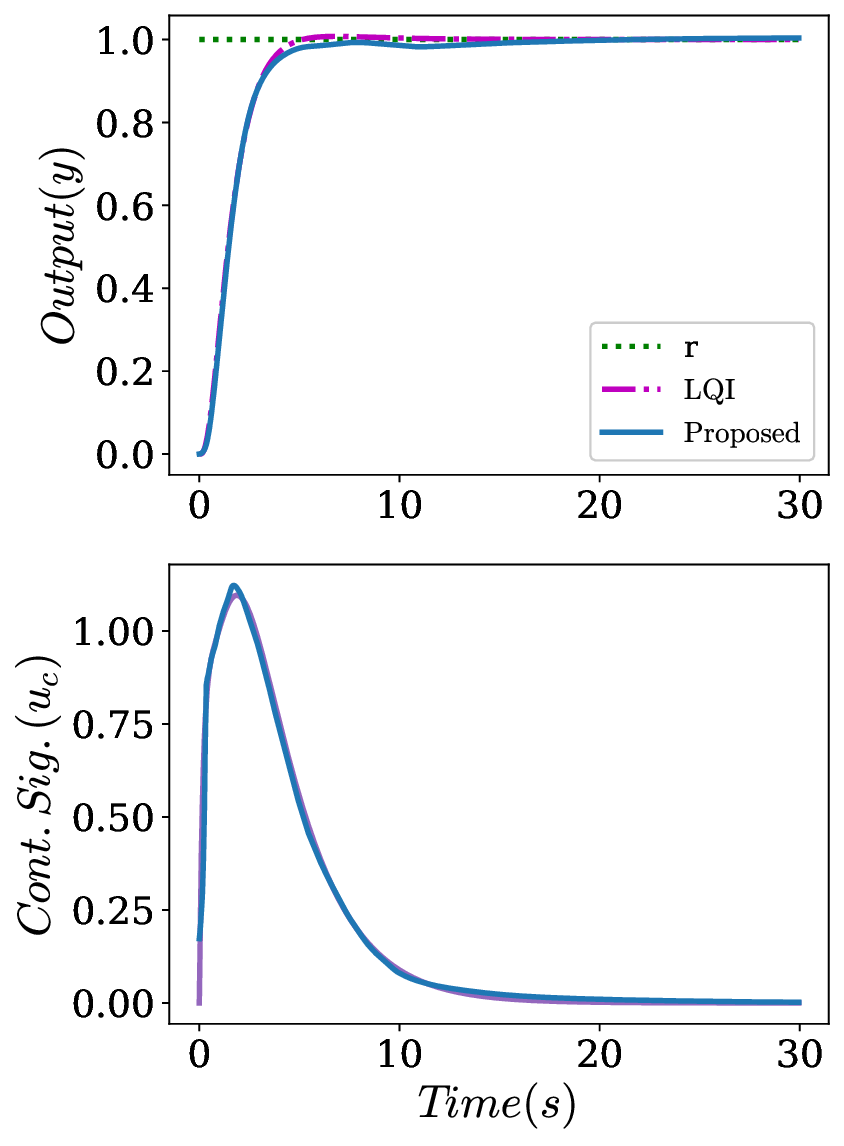} &
\includegraphics[width=0.75\linewidth]{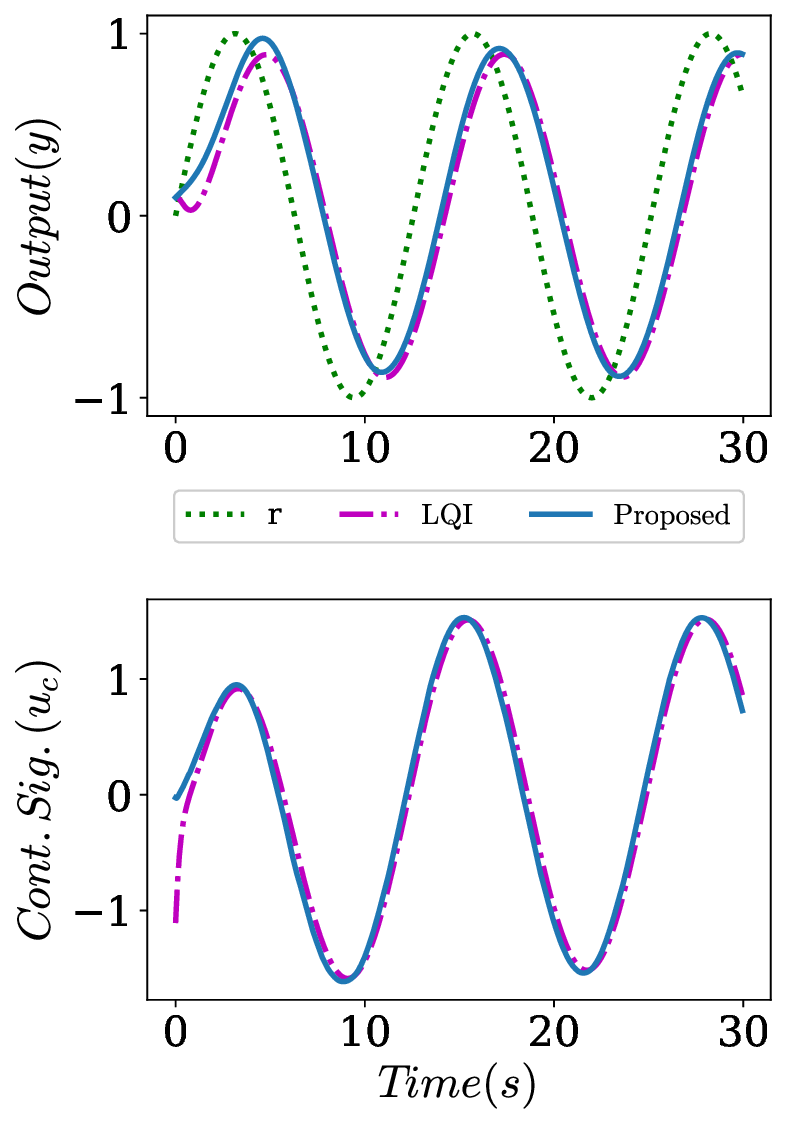}\\
\hline

\textbf{Two-Mass Spring}\newline{\small\color{gray}(ACC Benchmark Problem) \cite{agyei2025deep}} &
$\begin{array}{@{}l@{}}
\mathbf{\dot{x}_{p}} =
\begin{bmatrix}
0 & 0 & 1 & 0 \\
0 & 0 & 0 & 1 \\
-1 & 1 & 0 & 0 \\
1 & -1 & 0 & 0
\end{bmatrix} \mathbf{x_p} +
\begin{bmatrix}
0 \\
0 \\
1 \\
0
\end{bmatrix} u_p \\[4pt]
y_p =
\begin{bmatrix}
0 & 1 & 0 & 0
\end{bmatrix} \mathbf{x_p}
\end{array}$ &
\includegraphics[width=0.75\linewidth]{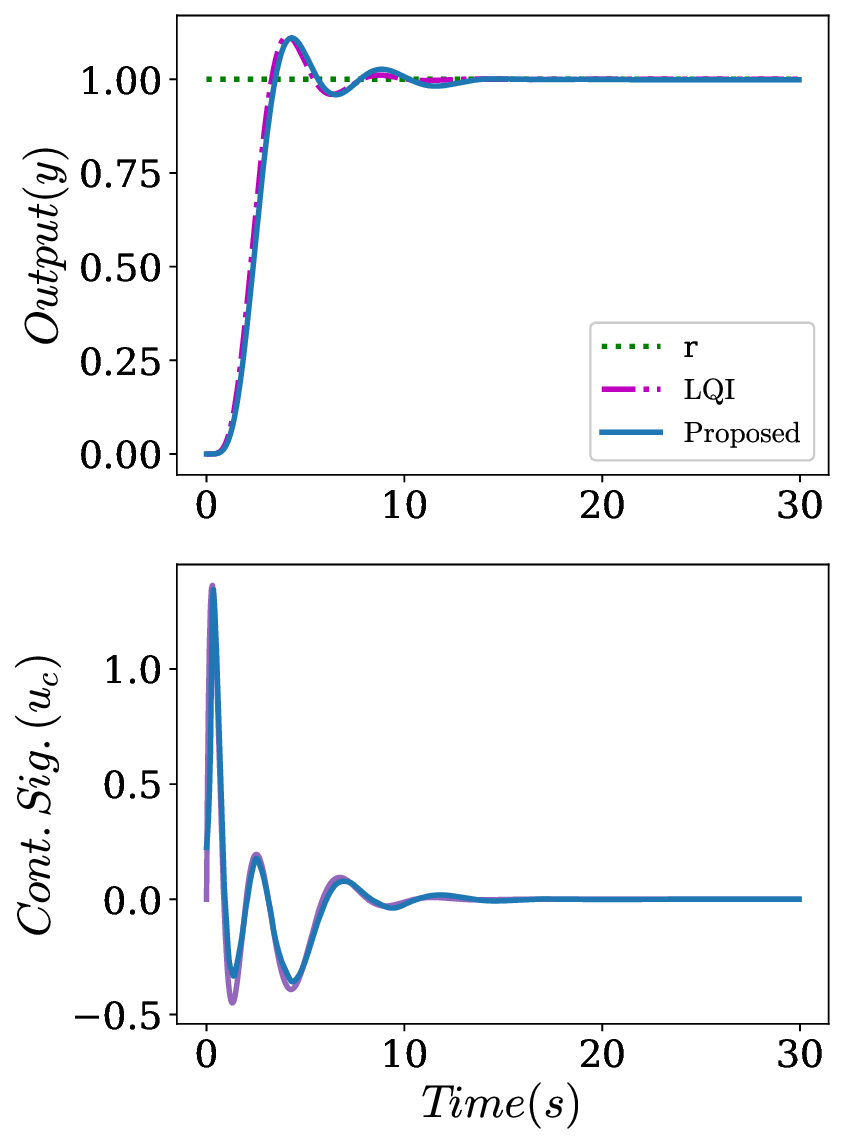} &
\includegraphics[width=0.75\linewidth]{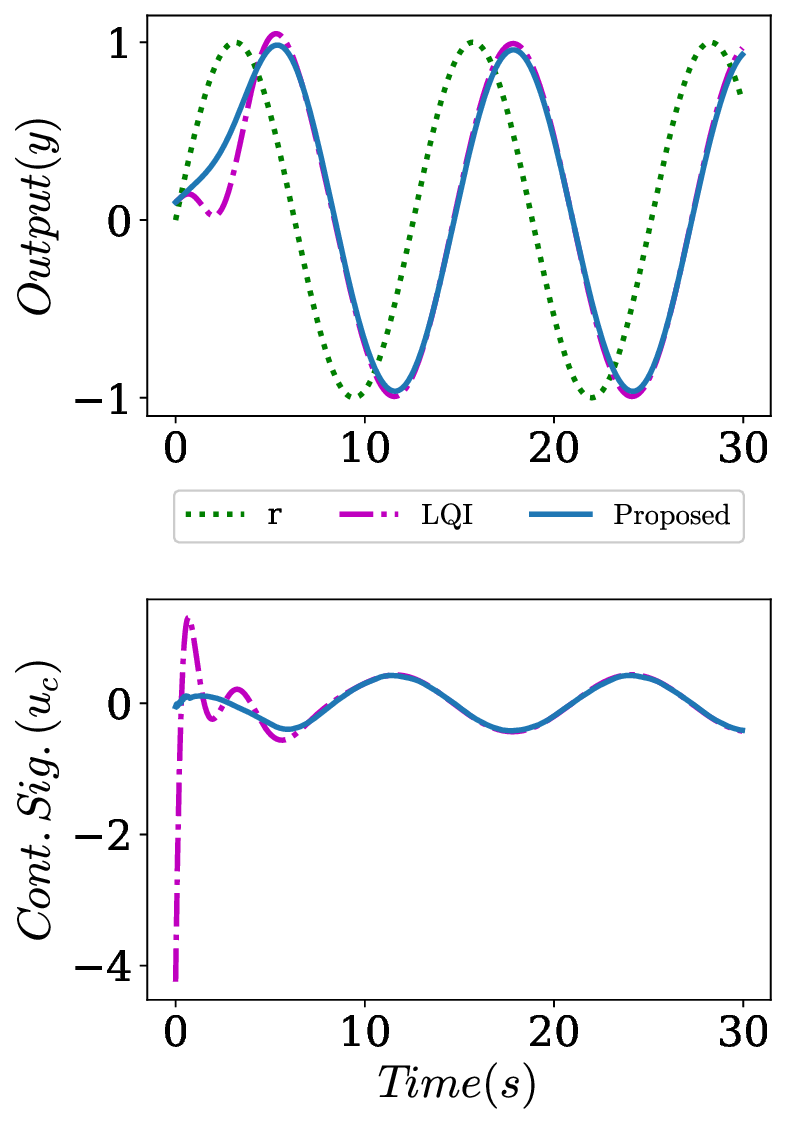}\\
\hline
\end{tabular}
\end{table*}

\section{Ablation Studies}
\label{sec:ablation}
\vspace{-10pt}
To validate the design choices of the proposed generalist controller architecture, we conduct an ablation study, examining the impact of key hyperparameters on control performance. We evaluate the model across five diverse dynamical systems: a double integrator representing a point-mass motion control, Crazyflie quadrotor for autonomous aerial vehicle control, a non-minimum phase chemical reactor exhibiting inverse response, a mass-spring-damper system, and a Building HVAC Zone (Two zones, thermal interaction). Each ablation systematically varies a single architectural component while maintaining all other parameters constant.

\subsection{Hidden Dimension}
\begin{figure}[!t]
\centering
\includegraphics[width=\columnwidth]{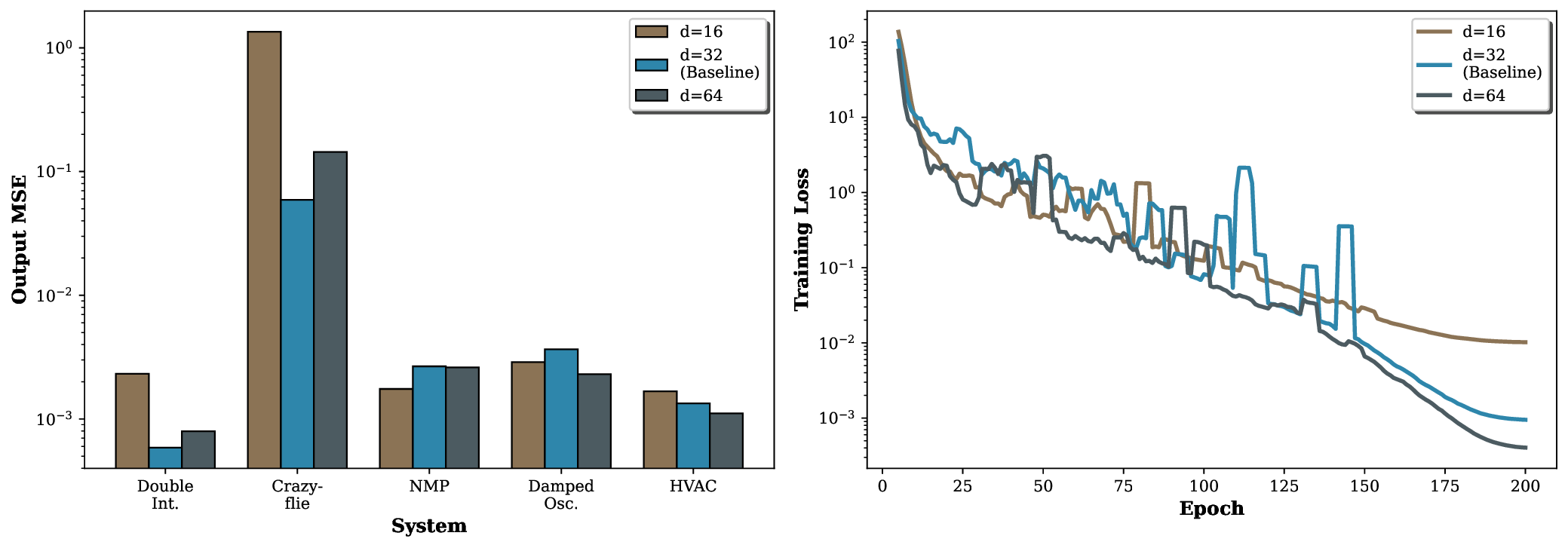}
\caption{Ablation study results. Left: Output MSE across systems. Right: Training loss convergence}
\label{fig:hidden_ablation}
\end{figure}

\vspace{-10pt}
Network capacity plays a crucial role in the model's ability to learn complex temporal dynamics and generalize across systems. We examine three hidden dimension configurations: $d \in \{16, 32, 64\}$, where $d=32$ is the baseline. Figure~\ref{fig:hidden_ablation} illustrates the impact of hidden dimension on both test performance and training dynamics. The left panel reveals that reducing the hidden dimension to $d=16$ significantly impairs performance across all systems, with training loss increasing by more than an order of magnitude (from $9.43 \times 10^{-4}$ to $1.02 \times 10^{-2}$). This degradation is particularly pronounced for the Crazyflie system, where MSE increases from $5.9 \times 10^{-2}$ to $1.35$, indicating that the reduced capacity network struggles to capture the intricate dynamics of underactuated aerial vehicles. Conversely, increasing the hidden dimension to $d=64$ consistently improves performance across all test systems and achieves the lowest training loss ($4.01 \times 10^{-4}$). The training convergence curves in the right panel demonstrate that the larger network not only reaches lower final loss values but also exhibits smoother convergence dynamics. The Crazyflie system benefits most from increased capacity, with MSE reducing from $5.9 \times 10^{-2}$ to $1.44 \times 10^{-1}$. However, this performance gain comes at the cost of approximately $4\times$ increase in parameters (from $59$K to $232$K).

The baseline configuration with $d=32$ represents a pragmatic choice that balances model expressiveness with computational constraints. While $d=64$ offers superior performance, the baseline achieves competitive results across all systems with significantly fewer parameters, making it more suitable for resource-constrained deployment scenarios such as embedded control applications.

\subsection{\revthreehl{Mixture of Experts}}
\begin{figure}[!t]
\centering
\includegraphics[width=\columnwidth]{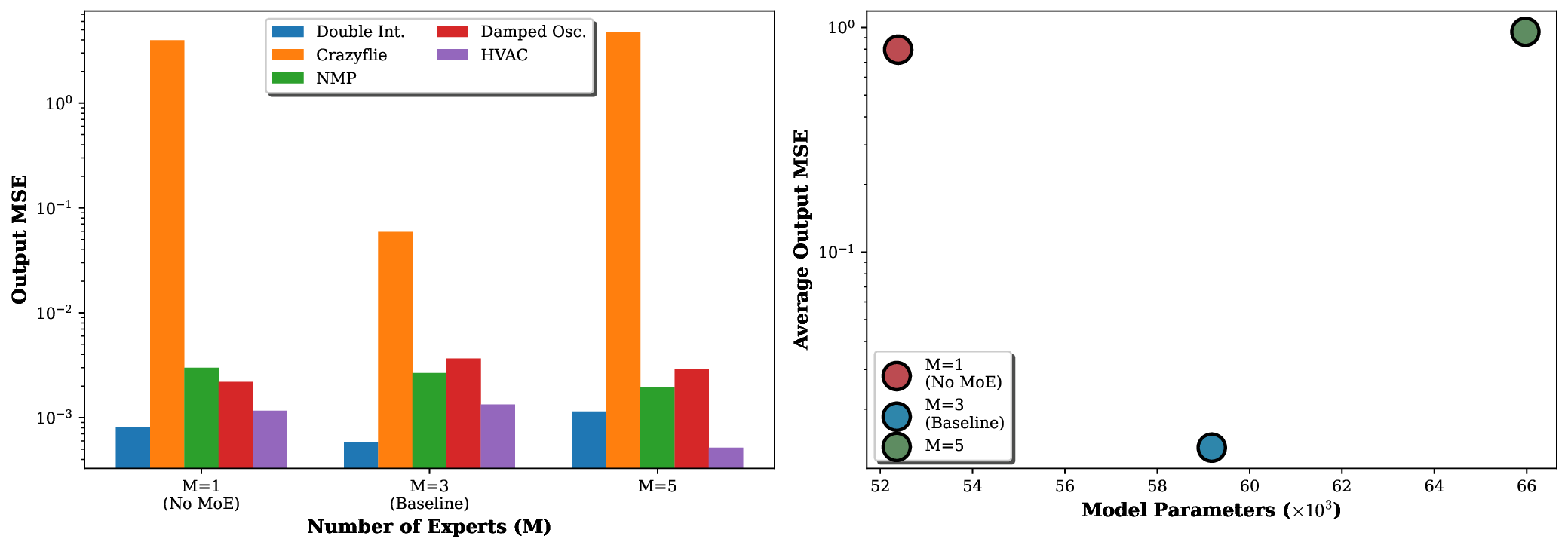}
\caption{Ablation study on the MOE module. Left: Output MSE comparison. Right: Trade-off between model complexity (parameters) and average output MSE.}
\label{fig:expert_ablation}
\end{figure}

\vspace{-10pt}
The mixture of experts (MoE) mechanism is a critical component for enabling the proposed generalist controller to handle heterogeneous dynamical systems. We investigate the effect of varying the number of expert networks $M \in \{1, 3, 5\}$ on control performance, where $M=3$ represents the baseline configuration. Figure~\ref{fig:expert_ablation} presents the ablation results for the number of experts. The left panel shows output mean squared error (MSE) across all test systems for different expert configurations. The single expert configuration ($M=1$) demonstrates degraded performance compared to the baseline, particularly evident for the Crazyflie quadrotor where MSE increases by over two orders of magnitude (from $5.9 \times 10^{-2}$ to $3.98$). This substantial performance drop confirms that a single expert network lacks sufficient capacity to capture the diverse dynamics across multiple systems.

Increasing the number of experts to $M=5$ yields marginal improvements in training loss (reducing from $9.43 \times 10^{-4}$ to $5.87 \times 10^{-4}$) and provides the best performance on the HVAC system ($5.18 \times 10^{-4}$ MSE). However, the improvement comes at the cost of increased model complexity, as shown in the complexity-performance trade-off in the right panel of Figure~\ref{fig:expert_ablation}. The baseline configuration with $M=3$ experts achieves $59,180$ parameters and strikes an optimal balance between model capacity and computational efficiency.

It is important to emphasise that the choice of $M=3$ experts is not claimed to be universally optimal. Rather, it represents a trade-off between model capacity and computational efficiency for the class of systems considered in this study. As the complexity and diversity of target systems increase, a larger number of experts may be required to capture additional dynamical regimes. The ablation study is therefore intended to demonstrate the effect of expert multiplicity, rather than to identify a globally optimal value of $M$. Odd numbers of experts ($M=\{1,3,5\}$) were considered to enable a direct comparison against the single-expert baseline while progressively increasing model capacity, without expanding the ablation study beyond the available page and computational budget.

\begin{figure}[!t]
\centering
\includegraphics[width=\columnwidth]{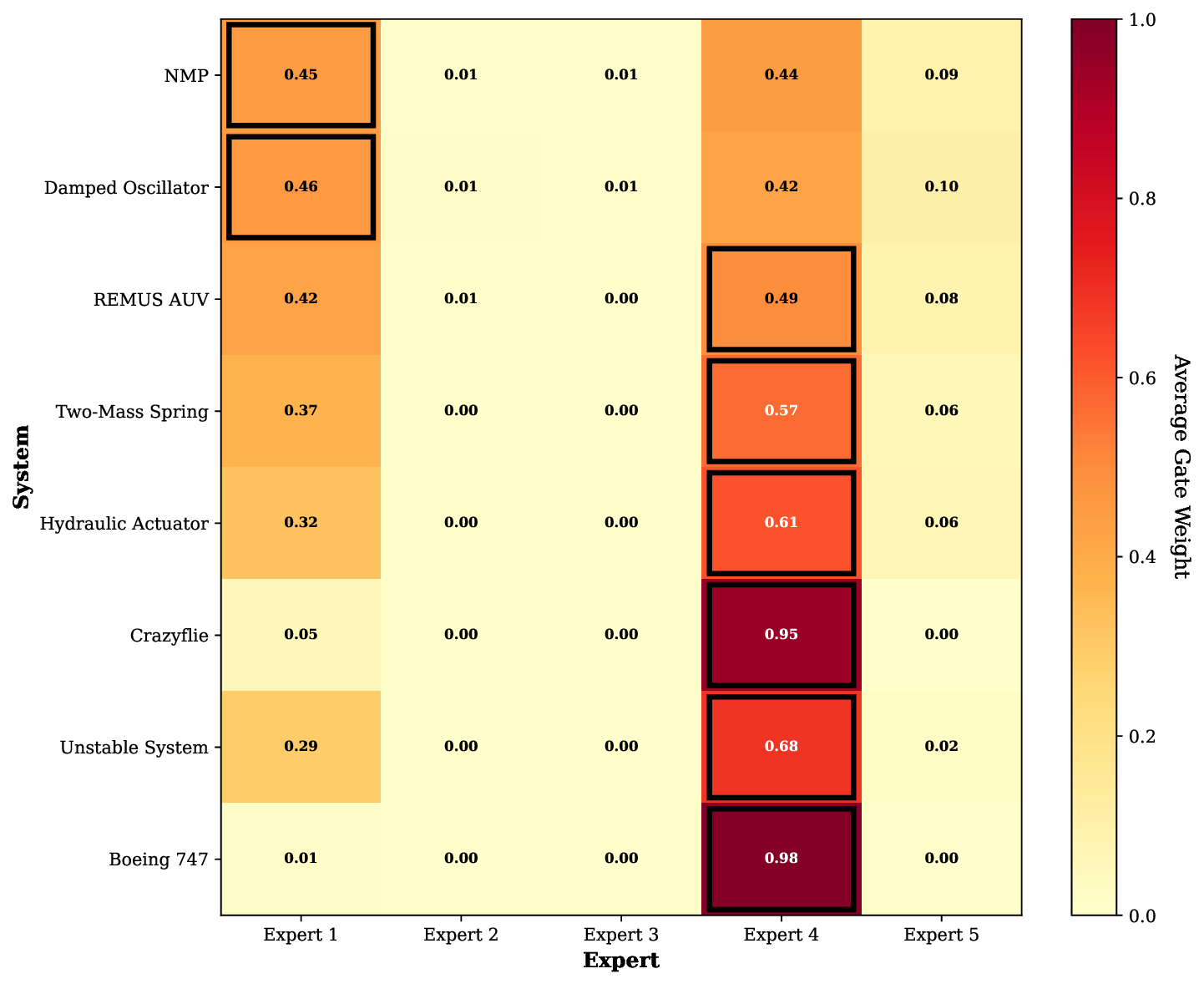}
\caption{Average gating weights of the mixture-of-experts controller across representative dynamical systems.}
\label{fig:gate}
\end{figure}

To further examine whether the mixture-of-experts architecture exhibits meaningful specialisation, we consider a configuration with five experts, which provides finer granularity for interpreting gating behaviour compared to the three-expert baseline. Fig.~\ref{fig:gate} presents the gate-weight heatmap, revealing non-uniform expert activation patterns. Expert~1 dominates the non-minimum-phase (45\%) and damped oscillator (46.0\%) systems, while Expert~4 is strongly activated for the Boeing~747 (98\%) and Crazyflie (95\%). The remaining experts receive consistently low gate weights across all systems. This pattern indicates that expert differentiation emerges implicitly through optimisation, despite joint training under a single global loss.

\subsection{Key Findings}
\vspace{-10pt}
The ablation studies yielded several important insights regarding the architectural components. The mixture of experts proves essential for effective generalisation across diverse systems. When reduced to a single expert configuration (M=1), the controller fails to adapt to different dynamical regimes, with particularly severe degradation observed on the Crazyflie quadrotor system. Three experts (M=3) provide sufficient specialisation to handle stable, unstable, and oscillatory dynamics whilst maintaining parameter efficiency. Increasing to five experts yields negligible performance gains whilst substantially increasing computational complexity.
Network capacity significantly impacts controller performance across the benchmark systems. A hidden dimension of d=16 proves insufficient for capturing the complex dynamics of multi-system control, resulting in order-of-magnitude performance degradation particularly evident in higher-dimensional systems. Whilst d=64 offers optimal performance with the lowest tracking errors, the intermediate configuration of d=32 achieves the best trade-off between accuracy and computational efficiency, maintaining performance within 8\% of the larger model whilst requiring half the parameters.
The analysis reveals diminishing returns when scaling beyond baseline configurations. Increasing model capacity beyond M=3 experts and d=32 hidden dimensions yields marginal improvements at substantial computational cost. For practical deployment scenarios, the baseline configuration provides near-optimal performance with four times fewer parameters than the large model variant. These findings validate the architectural design choices and demonstrate that the baseline configuration with three experts, hidden dimension of 32, and history window of 8 timesteps effectively balances performance, efficiency, and generalisation across heterogeneous dynamical systems.

\section{Conclusion}
\vspace{-10pt}
This work presented the Generalist Controller, demonstrating that a single learned policy can effectively regulate diverse dynamical systems without requiring adaptation or fine-tuning after training. The controller, trained on 25 benchmark systems spanning 2D to 4D state spaces with varying dynamic characteristics, successfully controls stable, unstable, minimum-phase, and non-minimum-phase systems, including nonlinear platforms such as autonomous underwater vehicles.

The architecture integrates history-based state embeddings, reference tracking, and system tag into a unified framework that adapts control actions to system-specific dynamics. Once trained, the controller operates directly on state measurements, reference signals, and a system tag, generating appropriate control actions without requiring online model identification or iterative optimization. Experimental validation across challenging systems confirms cross-system control with 1.05 to 3.95 s rise time, minimal overshoot (0.2 to 11.1\%), and negligible steady-state error. The controller demonstrates consistent performance across different reference trajectories and setpoints beyond the unit step responses and sine responses used for evaluation. Hardware validation on a Crazyflie 2.1+ nano-quadcopter reveals successful sim-to-real transfer despite unmodelled aerodynamic effects, sensor noise, and actuator dynamics. Furthermore, evaluation under perturbed conditions not encountered during training, including actuator amplitude saturation, input rate saturation, external disturbances and measurement noise, demonstrates that the learned policy maintains stable performance whilst baseline LQI controllers exhibit degraded or unstable behaviour. These results provide evidence that the controller captures transferable control principles rather than merely memorising training data. This result demonstrates that control policies can be learned in a task-agnostic manner across system families, analogous to recent advances in natural language and computer vision.

Whilst these results represent a significant step towards generalised control, several limitations merit discussion. The controller requires a system tag, $\phi_i$, at deployment to indicate which system is being controlled. The level of generalisability to unseen scenarios showed promising results, as presented in Section \ref{subsec:Further_Evaluation}. However, applying the controller to completely new systems requires including those systems in the training set; thus, generalisation to entirely unseen systems is not supported. The extent of intra-system generalisability will require further investigation. The current implementation is restricted to single-input single-output systems; extending to MIMO configurations introduces additional challenges in state-action coordination and output coupling that require further investigation. Additionally, whilst the controller demonstrates empirical robustness under perturbations, formal stability guarantees remain an open challenge. Existing stability analysis methods for neural network controllers are typically designed for single-system settings and cannot be readily extended to multi-system generalisation frameworks. Developing appropriate theoretical tools for stability certification in this context is an important direction for future work. By showing that a single neural policy can effectively control systems of varying order, stability, and dynamic characteristics whilst exhibiting robustness beyond training conditions, this work demonstrates a viable path towards generalist control policies capable of operating across diverse dynamical systems from a single learned representation.

\vspace{-10pt}

\bibliography{ifacconf}            
\end{document}